\documentclass{article}

\PassOptionsToPackage{numbers, compress}{natbib} 

\usepackage[preprint]{neurips_2025}

\usepackage[utf8]{inputenc} 
\usepackage[T1]{fontenc}    
\usepackage{hyperref}       
\usepackage{url}            
\usepackage{booktabs}       
\usepackage{amsfonts}       
\usepackage{nicefrac}       
\usepackage{microtype}      
\usepackage{xcolor}         

\usepackage{verbatim}
\usepackage{graphicx}
\usepackage{multirow}
\usepackage{algorithm}
\usepackage{algpseudocode}
\usepackage{amsthm}
\usepackage[most]{tcolorbox}
\usepackage{subcaption} 
\usepackage{tabularx} 
\usepackage{adjustbox}
\usepackage{multicol, multirow, makecell}
\usepackage{wrapfig}
\usepackage{xspace}

\definecolor{promptcolor}{RGB}{0,0,255} 
\definecolor{responsecolor}{RGB}{0,128,0} 

\newcommand{\x}{\mathbf{x}}

\newcommand{\K}{\mathbf{K}} 
 
\newcommand{\bk}{\mathbf{k}}

\newcommand{\q}{\mathbf{q}}

\newcommand{\V}{\mathbf{V}}

\newcommand{\bv}{\mathbf{v}}

\newcommand{\method}{SiftAttention\xspace}

\newcommand{\prompt}[1]{%
    \begin{tcolorbox}[colframe=promptcolor, colback=promptcolor!10, sharp corners, boxrule=0.5pt, arc=2mm]
        \begin{tcolorbox}[colframe=promptcolor, colback=white, rounded corners, boxrule=0.5pt, arc=2mm]
            \centering\small\textbf{Prompt}
        \end{tcolorbox}
        #1
    \end{tcolorbox}
}

\newcommand{\response}[2]{%
    \begin{tcolorbox}[colframe=responsecolor, colback=responsecolor!10, sharp corners, boxrule=0.5pt, arc=2mm]
        \begin{tcolorbox}[colframe=responsecolor, colback=white, rounded corners, boxrule=0.5pt, arc=2mm]
            \centering\small\textbf{#2}
        \end{tcolorbox}
        #1
    \end{tcolorbox}
}

\newtcolorbox{researchquestion}{
    colback=blue!5!white,
    colframe=blue!75!black,
    boxrule=0mm,
    arc=0mm,
    left=10pt,
    right=10pt,
    boxsep=1pt,
    toptitle=1pt,
    bottomtitle=1pt,
}

\title{Power Law Guided Dynamic Sifting for Efficient Attention}

 \author{%
   Nirav Koley\thanks{Equal contribution},\,\,
   Prajwal Singhania\footnotemark[1],\,\,
   Abhinav Bhatele
   \AND
   Department of Computer Science, University of Maryland\\
   College Park, MD 20742\\
   \texttt{prajwal@umd.edu},
   \texttt{bhatele@cs.umd.edu}
 }

\begin{document}

\maketitle

\begin{abstract}
Efficient inference on GPUs using large language models remains challenging due
to memory bandwidth limitations, particularly during data transfers between
High Bandwidth Memory (HBM) and SRAM in attention computations. Approximate
attention methods address this issue by reducing computational and memory
overhead but often rely on expensive top-$k$ operations, which perform poorly
on GPUs. We propose \method, a novel approximate attention method that replaces
the top-$k$ step with a computationally efficient element-wise filtering 
operation based on a threshold value. Our intuition for doing this is based on our empirical
observation that the $\tau$-th quantile of attention scores follows a
predictable power-law over sequential generation steps. Exploiting this
insight, our approach dynamically estimates a threshold value per prompt at each
generation step.  Only attention scores above this threshold and their
corresponding value vectors are loaded/used to compute the attention output,
reducing data movement between HBM and SRAM. Our evaluation demonstrates that
\method preserves model quality better than existing approximate attention
methods while reducing memory bandwidth usage when loading value vectors. 


\end{abstract}

\section{Introduction}
\label{sec:intro}
Large language models (LLMs) have transformed natural language processing and
many other areas in computer science, achieving state-of-the-art results across
a wide range of tasks. Powered by the transformer
architecture~\citep{vaswani2017attention}, these models leverage the
self-attention mechanism~\citep{bahdanau2014neural} to learn complex language
patterns. As the sizes of large language models (LLMs) have grown from millions
to hundreds of billions of parameters, their computational, memory, and
consequently energy demands have also increased significantly.

The self-attention mechanism, originally applied to relatively small models, now
poses a major efficiency bottleneck, especially during auto-regressive
generation for long sequence lengths. At each generation step, the attention
mechanism computes a weighted sum over all previous token states stored in a
key-value (KV) cache. Transferring this data from GPU High Bandwidth Memory
(HBM) to on-chip SRAM is costly and leads to significant memory bandwidth
bottlenecks~\citep{ivanov2021data}.  Additionally, it requires dot-product
computation between the current input and all previous token states, leading to
quadratic computational complexity with respect to sequence length.

\begin{figure}[ht]
    \centering
    \includegraphics[width=1\textwidth]{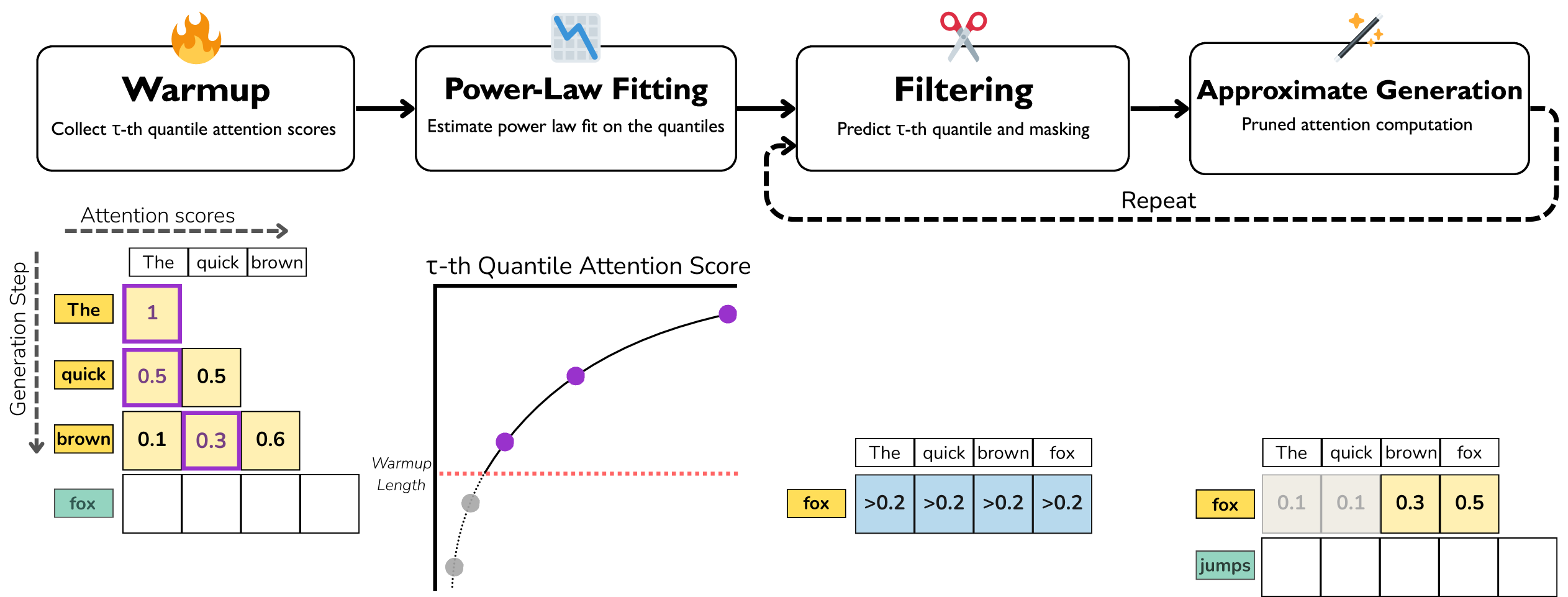}
    \caption{Overview of \method}
    \label{fig:overview}
\end{figure}

Several methods have been proposed to address the computational challenges
associated with the attention mechanism. Our work focuses on a particular class
of methods that aim to approximate the attention computation by reducing the
number of tokens used in the attention computation, trading off accuracy for
efficiency.  These methods either prune the KV-cache
permanently~\citep{zhang2023hH2O} for memory savings or dynamically select a
subset of tokens per generation step~\citep{ribar2023sparq, lee2024infinigen,
singhania:neurips2024} to reduce the amount of key-value pairs to attend to. The
latter set of methods typically use a top-$k$ selection operation to select the
top $k$ tokens with the highest attention scores, which is inefficient on GPUs
due to thread-synchronization overheads. We introduce \emph{\method}, a new
approximation strategy that replaces top-$k$ with an element-wise filtering
operation. 

Our approach is based on a key empirical finding: the $\tau$-th quantile of
attention scores follow a predictable power-law decay over generation steps.
This trend is consistent across models and datasets and enables us to estimate
quantile thresholds dynamically using a small warmup window. 
Based on this insight, we design \method with two-phases: a short \emph{warmup},
and an \emph{approximate generation} phase. During the warmup, seen on the left
of Figure~\ref{fig:overview}, we record the $\tau$-th quantile of attention
scores at each generation step and fit a power-law curve to these samples.
During approximate generation, seen on the right of Figure~\ref{fig:overview},
we use our fitted power-law to predict the $\tau$-th quantile of attention
scores at each generation step, retaining only those keys whose attention
weights are above this threshold and loading their corresponding value vectors
to SRAM. 

We evaluate \method on perplexity, short-context, and long-context generation
tasks across multiple models. Our results demonstrate that \method maintains a
high level of model quality, with negligible degradation in perplexity (within
0.1) and task-specific metrics. We also implement a fused
Triton~\citep{triton:openai} kernel for \method to demonstrate that our method
reduces data movement in the GPU memory hierarchy when loading value vectors,
similar to other top-$k$ methods.

In summary, our contributions are as follows:
\begin{itemize}
    \item We discover and analyze that attention score quantiles follow a
    power-law trend across generation steps and demonstrate that this can be
    estimated using a few warmup steps.
    \item We propose \method, an approximate attention method that avoids the
    expensive top-$k$ sorting operation by compute-efficient filtering using a
    quantile-based threshold value.
    \item We demonstrate \method's effectiveness across multiple benchmarks,
    including perplexity, short-context, and long-context generation tasks, and
    showing that it maintains model quality better than existing approximate
    attention methods.
\end{itemize}

\section{Background and Related Work}
\label{sec:bg}
\textbf{Auto-regressive Decoding:} During auto-regressive decoding, transformers
generate one token at a time by applying causal self-attention over all
previously generated tokens~\citep{vaswani2017attention}. At generation step
$S$, the attention output is computed as:
\begin{equation}
  \mathbf{o}_{S} = \mathrm{softmax}\left( \frac{\mathbf{q}_{S}\K_{:S}^{T}}{\sqrt{D}} \right) \V_{:S}
\end{equation}
where $\mathbf{q}_{S} \in \mathbb{R}^{1\times D} $ is the query, and
$\mathbf{K}_{:S}, \mathbf{V}_{:S} \in \mathbb{R}^{S \times D}$ are the key and
value caches, respectively. 

\textbf{Power-Law Decay:} A power-law decay describes relationships of the form:
$y = \alpha \cdot x^{-\beta}$, where $\alpha, \beta \in \mathbb{R}$ and $\alpha,
\beta > 0$. Assuming a multiplicative error in $y$, this can be linearized in
log-log space to enable efficient parameter estimation via linear regression,
as:
\begin{equation}
    \log y = \log \alpha - \beta \log x
\end{equation}
\subsection{Related Work}
Empirical analyses of pre-trained transformers show that only a small subset of
past keys receive the majority of the attention weight~\citep{nahshan2023linear}.
This has motivated a range of approximate attention methods that exploit this
sparsity to either reduce the cost of full attention~\citep{topkattention,
ribar2023sparq, singhania:neurips2024, lee2024infinigen} or prune unimportant
key-value pairs to reduce memory overhead~\citep{liu2023scissorhands,
zhang2023hH2O}. While traditional methods focus on patterns within a single
generation step, very few methods explore how attention scores evolve across
generation steps. To the best of our knowledge, our work is the first to
empirically demonstrate that attention score quantiles exhibit a predictable
power-law decay over time and to then use this insight to sparsify attention
scores dynamically during inference.

The original Top‑$k$ method~\citep{topkattention} selects the most relevant $k$
keys via a global sort over exact attention scores. Despite impressive
downstream performance, the top‑$k$ operation introduces a full‑row dependency
that limits parallel execution on modern GPUs. While recent methods like
Loki~\citep{singhania:neurips2024} and SparQ~\citep{ribar2023sparq} aim to
approximate attention scores in low-dimensional subspaces, they still rely on
the top‑$k$ operation to select the most important scores, which remains a
bottleneck~\citep{zhang2023parallel}. \method aims to replaces the top-$k$
operation with a simple element-wise threshold comparison, making it highly
parallelizable on GPUs.

Token-eviction methods, such as Scissorhands~\citep{liu2023scissorhands} and the
H$_2$O~\citep{zhang2023hH2O}, reduce memory usage by deleting tokens
permanently, which can lead to a significant degradation in model quality. In
contrast, \method retains the full KV-cache and sparsifies only during attention
computation, prioritizing task performance over memory savings.

The most closely related works to \method are LeOPArd~\citep{li2022accelerating}
and Top-$\theta$ Attention~\citep{berestizshevsky2025top}. LeOPArd learns fixed
thresholds per transformer layer during training and drops scores below them at
inference time, achieving 1.9$\times$ speedups on specialized hardware. However,
it introduces training/fine-tuning overhead and lacks flexibility to adapt
thresholds based on sequence specific dynamics. Top-$\theta$ estimates
thresholds per head, layer, and generation step using an offline calibration set
with some limited online fine-tuning. This requires storing calibrated
thresholds for all generation steps, which can be memory-intensive for long
contexts. \method is the first fully online, dataset-agnostic, and
model-agnostic sparsification technique that doesn't require any offline step
which can dynamically adapt to sequence/query dynamics. 

\section{Power-Law Analysis}
\label{sec:powerlawanaylis}
\label{sec:intro}
In this section, we investigate the evolution of attention score quantiles over
sequential generation steps and discover a consistent empirical pattern: the
$\tau$-th quantiles exhibit a decay that closely follows a power-law. We then
show the efficacy of capturing this trend using a power-law fit and introduce a
method to estimate the parameters of this fit with comparatively minimal
overhead.

\subsection{Analyzing the Temporal Decay of Attention Score Quantiles}
\label{subsec:powerlaw}
To examine whether attention scores exhibit consistent and predictable behavior
over generation steps, we analyze them across multiple models on a perplexity
evaluation task. Specifically, we focus on specific quantiles of the attention
scores and fit power-law models to capture their decay over time.

\textbf{Experimental Setup:} We evaluate five models spanning a diverse range of
architectures and scales: Llama-3.1 (8B Instruct and 70B Instruct), Llama-3.2 3B
Instruct, Mistral 7B, Qwen2.5 7B~\citep{grattafiori2024llama3herdmodels,
jiang2023mistral, yang2024qwen2}. Evaluations are conducted on WikiText-2 (test
split)~\citep{wikitext-103} and C4 (custom split)
~\citep{raffel2023exploring}, using a perplexity evaluation pipeline. Each model
is run on sequences of length 4096 tokens. The experiments are run on Nvidia
GH200 and H100 GPUS, with larger models distributed across multiple GPUs using
AxoNN~\citep{singh:ipdps2022, singh:arxiv2024}.

We track the attention scores at selected quantile levels $\tau$: 0.5, 0.75, and
0.875, denoting the $\tau$-th quantile at generation step $i$ as $\theta_{i,
\tau}$. For each prompt, layer, and head, we extract the quantile time series
and fit a power-law curve of the form $\hat{\theta}_{i, \tau} \approx \alpha
\cdot i^{-\beta}$. We fit the power-law by performing linear regression in
log-log space, which enables both efficient closed-form estimation using
standard tensor operations and implicitly models multiplicative noise. To
evaluate the quality of the fit, we compute the coefficient of determination
($R^2$), defined as:
\begin{equation}
    R^2 = 1 - \frac{\sum_i (\ell_i - \hat{\ell}_i)^2}{\sum_i (\ell_i - \bar{\ell})^2}
\end{equation}
where $\ell_i$ is the logarithm of the $\tau$-th quantile score at generation
step $i$, $\ell_i = \log(\theta_{i, \tau})$, $\hat{\ell}_i$ is the logarithm of
the predicted value from the power-law fit, and $\bar{\ell}$ denotes the mean
value of $\ell_i$. This metric is computed for each prompt, layer, and head. 

\begin{figure}[h]
  \centering
    \includegraphics[height=1.31in]{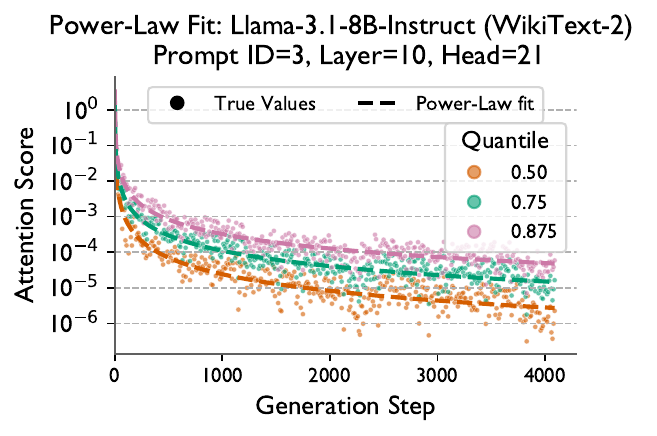}
    \includegraphics[height=1.31in]{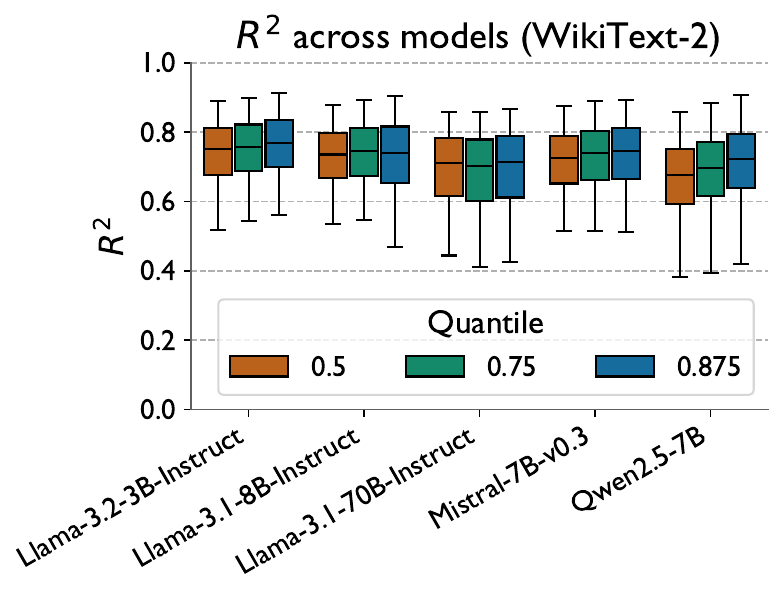}
    \includegraphics[height=1.31in]{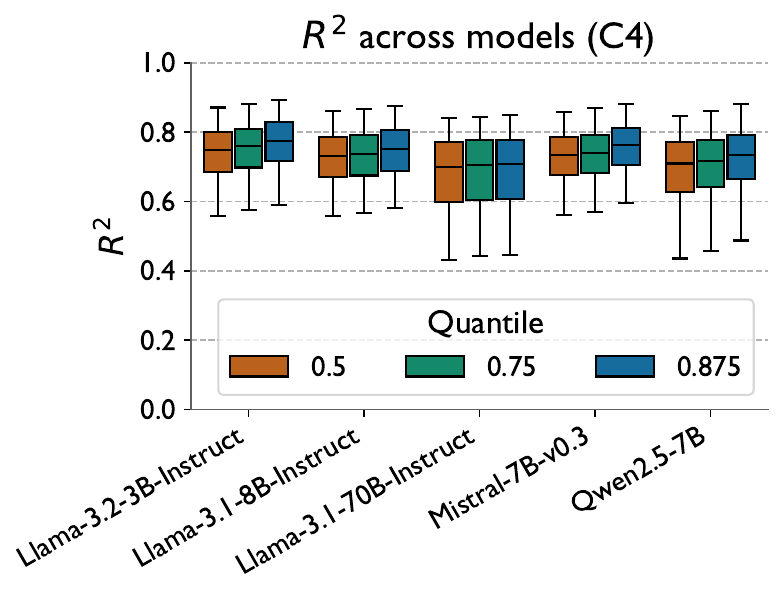}
    \caption{Power-Law fit for attention score quantiles (log y-scale) over
    generation steps for Llama-3.1 8B on WikiText-2 (left). Distribution of
    $R^2$ values for power-law fits across prompts, layers, and heads for
    various models on WikiText-2 (middle) and C4 (right) datasets. Boxes denote
    the median, 25$^{th}$ and 75$^{th}$ percentiles, and the whiskers denote the
    5$^{th}$ and 95$^{th}$ percentile values.}
    \label{fig:powerlaw-fit}
\end{figure}

\textbf{Analysis:} We begin by analyzing the temporal trend in attention score
quantiles for Llama-3.1 8B on WikiText-2 (Figure~\ref{fig:powerlaw-fit}, left).
The plot depicts the evolution of the 0.5, 0.75, and 0.875 quantiles over
generation steps for a specific prompt, layer, and head index. Each line
reflects a power-law fit, plotted alongside the true quantile values. For visual
clarity, the generation steps are down-sampled to onlyshow every 10$^{th}$ step.
Despite some local noise, we observe a remarkably consistent decline in quantile
values over time, with high correspondence to the fitted power-law curves.
Similar trends for other models can be found in appendix
\ref{sec:extended-power-law-analysis}.

To demonstrate the generalizability of this observed trend, we report the $R^2$
values for power-law fits aggregated across all layers, heads, and prompts for
different models (Figure~\ref{fig:powerlaw-fit}, middle and right). Across
models of different architectures and sizes, the power-law fit consistently
achieves high $R^2$ values, with median values between 0.6 and 0.8. Although a
small fraction of fits exhibit lower quality—particularly at the 5th percentile,
which can fall to around 0.4 - the overall distribution of $R^2$ indicates that
the power-law fit is a good approximation for the decay of attention score
quantiles over generation steps. This trend further holds across both datasets
used. This consistent result across diverse settings supports the hypothesis
that the quantile decay might be a fundamental and architecture-agnostic
property of transformer models.

This insight has significant practical implications: it suggests that the
trajectory of attention score quantiles can be predicted reliably across
different models, potentially enabling efficient threshold estimation without
the need for an expensive top-$k$ operation at each step. Next, we explore the
possibility of estimating the power-law parameters using a small warmup phase. 

\subsection{Estimating Power-Law Fit Parameters with a Warmup Phase}
\begin{figure}[h]
  \centering
    \includegraphics[height=1.75in]{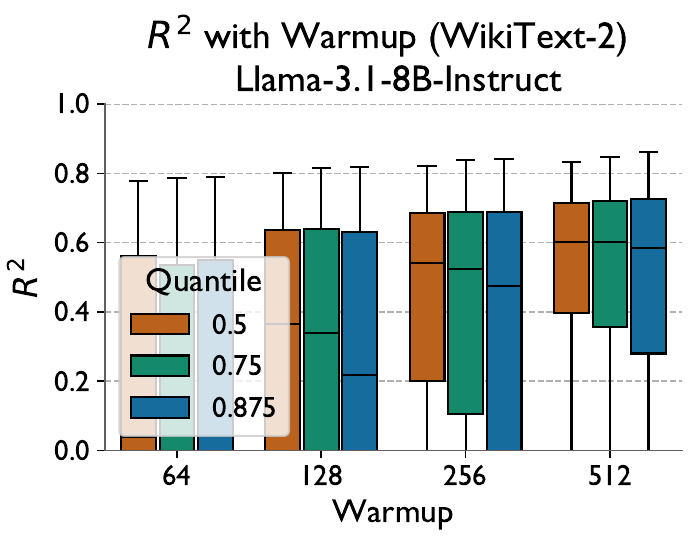}
    \includegraphics[height=1.75in]{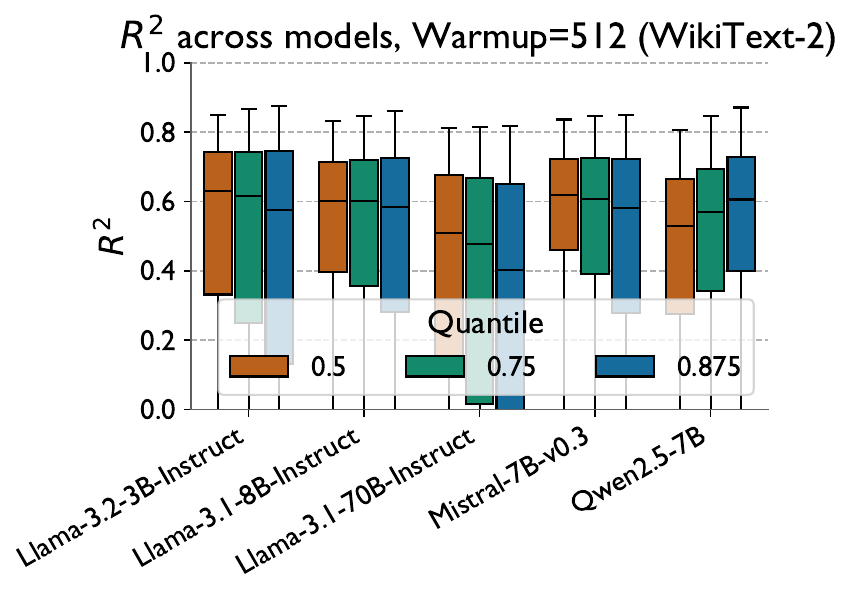}
    \caption{$R^2$ of power-law fits for attention score quantiles with warmup.
      Left:  $R^2$ distribution for Llama-3.1-8B-Instruct on WikiText-2,
      evaluated across warmup sizes of 64, 128, 256, and 512 steps. Right: Fit
      quality across models with a warmup of 512. Longer warmups consistently
      yield stronger fits across models.}
    \label{fig:powerlaw-fit-warmup}
\end{figure}

Having established that attention score quantiles follow an approximate
power-law decay, we now focus on estimating the parameters of this curve during
inference. We explore whether a small number of initial generation steps can be
used to estimate the power-law parameters accurately.

Following the experimental setup described in Section~\ref{subsec:powerlaw}, we
now fit the power-law function using only the values from the first $w$
generation steps - henceforth referred to as the warmup phase. We evaluate
four warmup lengths: $w$: 64, 128, 256, and 512. The power-law parameters are
estimated using only the warmup steps, but the resulting fit is evaluated across
all generation steps (including both warmup and post-warmup) by computing the
coefficient of determination ($R^2$). This warmup-based $R^2$ captures
out-of-sample generalization and as a result, $R^2$ values can be negative when
the power-law fit performs worse than simply predicting the mean of the
log-quantile values across the full sequence.

Figure~\ref{fig:powerlaw-fit-warmup} (left) shows how fit quality varies with
the number of warmup steps for Llama-3.1-8B-Instruct on WikiText-2. With short
warmup lengths (e.g., $w=$64), the $R^2$ values are often low, with a
significant fraction falling below zero (truncated in the plot to the range $[0,
1]$). As the warmup length increases, the fit quality improves steadily. The
same pattern can be seen across various models, the results of which are shown
in appendix \ref{sec:extended-power-law-analysis}. With $w=$512, the median
$R^2$ across all quantiles approaches or exceeds 0.6, indicating strong
generalization. Notably, 512 warmup steps still correspond to just 12.5\% of the
full 4096-token sequence, suggesting that accurate power-law parameter
estimation requires only a small fraction of the total generation steps. Later,
we will see that empirically, this is an adequate warmup length to achieve high
downstream performance.

\section{\method: Power-Law Guided Dynamic Filtering}
\label{sec:method}
In this section, we introduce \method, an approximate attention mechanism that
exploits the power-law behavior of attention score distributions. We describe
its two-phase design: a warmup phase for estimating power-law parameters,
followed by an approximate generation phase that uses these estimates to prune
attention computations efficiently. We also analyze the computational complexity
of \method and discuss its implementation in practical settings.

\subsection{Two-Phase Algorithm Design}
\method operates in two distinct phases: a \emph{warmup} phase
(Algorithm~\ref{alg:threshold-warmup}) and an \emph{approximate} generation
phase (Algorithm~\ref{alg:threshold-approx}).

\begin{algorithm}[h]
  \caption{\method: Warmup Phase}
  \label{alg:threshold-warmup}
  \renewcommand{\algorithmicrequire}{\textbf{Input:}}
  \begin{algorithmic}[1]
  \Require At the $S^{th}$ step - Input: $\x_S \in \mathbb{R}^{1 \times D}$,
    KV-cache: $\K_{:S-1}, \V_{:S-1} \in \mathbb{R}^{(S - 1) \times D}$, Warmup
    Steps $w$, Quantile Level $\tau$, Past Quantile Scores:
    $\boldsymbol{\theta}_{:S-1} \in \mathbb{R}^{(S - 1) \times 1}$ \Ensure $S
    \leq w$ 
    \State $\q_S, \bk_S, \bv_S \gets \textsc{ComputeQKV}(\x_S)$ \State $\K_{:S}
    \gets \textsc{append}(\K_{:S-1}, \bk_S), \V_{:S} \gets
    \textsc{append}(\V_{:S-1}, \bv_S)$ \State $\boldsymbol{a}_S \gets
    \textsc{softmax}\left(\frac{\q_S \K_{:S}^\top}{\sqrt{D}}\right)$ \State
    $\theta_S \gets \textsc{Quantile}(\boldsymbol{a}_S, \tau)$
    \Comment{$\tau$-th quantile of attention scores} \State
    $\boldsymbol{\theta}_{:S} \gets \textsc{Concat}(\boldsymbol{\theta}_{:S-1},
    \theta_S)$ \If{$S = w$} \Comment{Warmup complete} \State $\alpha, \beta
    \gets \textsc{FitPowerLaw}(\boldsymbol{\theta}_{:S})$ \State \Return
    $\boldsymbol{a}_S \V_{:S}, \alpha, \beta$ \Else \State \Return
    $\boldsymbol{a}_S \V_{:S}$ \EndIf
  \end{algorithmic}
\end{algorithm}

\begin{algorithm}[h]
  \caption{\method: Approximate Generation Phase}
  \label{alg:threshold-approx}
  \renewcommand{\algorithmicrequire}{\textbf{Input:}}
  \begin{algorithmic}[1]
  \Require At the $S^{th}$ step - Input: $\x_S \in \mathbb{R}^{1 \times D}$,
  KV-cache: $\K_{:S-1}, \V_{:S-1} \in \mathbb{R}^{(S - 1) \times D}$, Warmup
  Steps $w$, Powerlaw Fit Parameters - $\alpha, \beta$ 
  \Ensure $S > w$
  \State $\q_S, \bk_S, \bv_S \gets \textsc{ComputeQKV}(\x_S)$ 
  \State $\K_{:S} \gets \textsc{Append}(\K_{:S-1}, \bk_S),\quad \V_{:S} \gets \textsc{Append}(\V_{:S-1}, \bv_S)$ 
  \State $\boldsymbol{a}_S \gets \textsc{softmax}\left(\frac{\q_S \K_{:S}^\top}{\sqrt{D}}\right)$ 
  \State $\eta_S \gets \alpha \cdot S^{-\beta}$ \Comment{Power-Law Estimation of Quantile} 
  \State $\mathbf{indices} \gets \{i\ |\ \boldsymbol{a}_S[i] > \eta_S\}$ \Comment{Filtering} 
  \State $\boldsymbol{a}_S^{\prime} \gets \boldsymbol{a}_S[\mathbf{indices}], \quad \V_{:S}^{\prime} \gets \V_{:S}[\mathbf{indices}]$ \Comment{Retain only the indices above threshold}
  \State \Return $\boldsymbol{a}_S^{\prime}\V_{:S}^{\prime}$
  \end{algorithmic}
\end{algorithm}

\textbf{Warmup Phase:} During the warmup phase, we first compute the exact
attention scores using the dot product of query and key vectors in the KV-Cache
(Lines 1–2), followed by a softmax operation (Line 3). We then record the
$\tau$-th quantile of the computed attention scores for each prompt, layer, and
head (Lines 4–5). This step is repeated until the warmup phase ends ($S \le w$).

At the end of the warmup phase, we fit a power-law curve to these recorded
quantile values across generation steps to obtain parameters $\alpha$ and
$\beta$ (Line 7). The fitting is performed efficiently by applying a logarithmic
transform to both the generation step index $S$ and quantile values, followed by
linear regression in the subsequent log-log space. This method provides a
closed-form solution using standard tensor operations, eliminating iterative
optimization, and aligns naturally with the multiplicative noise assumption
inherent in modeling probabilities.

\textbf{Approximate Generation Phase.} Once the warmup phase ends and the
power-law parameters have been fit, the model enters the approximate generation
phase (Algorithm~\ref{alg:threshold-approx}). At each subsequent generation
step, we use the learned power-law model to estimate the $\tau$-th quantile of
attention scores based on the current step index (Line 4). This predicted
quantile serves as a dynamic threshold that adapts over time, reflecting the
prior empirical observations in attention score distributions.

Using this threshold, we prune the attention weights by zeroing out all values
below the predicted quantile (Line 5). Only the attention scores above the
threshold are retained, and the corresponding value vectors are extracted from
the KV-cache (Line 6). This selective sparsification reduces the number of
values that need to be loaded from high-bandwidth memory (HBM) into on-chip
SRAM, just as in top-$k$ methods. Finally, the attention output is computed
using only the surviving attention scores and their associated value vectors
(Lines 7). 

\subsection{Computational Cost Analysis}
We now present a run-time cost analysis of \method, compared to the Top-$k$
method. Let $S$ be the total number of generation steps (or tokens in the
KV-cache), $w$ the number of warmup steps.

The warmup phase of \method and the Top-$k$ method are almost identical, with
two main differences: \method requires an extra step of storing the $\tau$-th
quantile of attention scores at each step, and the final value projection uses
all the attention scores (Line 8 \& 10, Algorithm~\ref{alg:threshold-warmup})
(as opposed to Top-$k$ which uses a fixed $k$ scores). Given that the quantile
score can be directly stored in a buffer, without any additional copying, we can
approximate the difference in runtimes as:
\begin{equation}
T_{\text{\method, Warmup}} - T_{\text{Top-}k\text{ Attention}} \approx (T_{\text{proj-}V} - T_{\text{proj-}V’})
\end{equation}
where $T_{\text{proj-}V}$ is the runtime cost of computing the vector-matrix
product between the full attention scores and the value matrix, and
$T_{\text{proj-}V’}$ is the cost of the same operation between the pruned
attention scores and the value matrix.

During the approximate generation phase, \method thresholds the attention scores
before value projection, instead of a top-$k$ selection, and thus the difference
in runtime can be approximated as:
\begin{equation}
T_{\text{\method, ApproxGen}} - T_{\text{Top-}k\text{ Attention}} \approx T_{\text{threshold}} - T_{\text{top-}k}
\end{equation}
All the other operations are the same as in the Top-$k$ method. Based on these
approximations, we can compare the runtime of \method with the Top-$k$ method
over $S$ generation steps (with $w$ warmup steps), as follows:
\begin{equation}
T_{\text{\method}}^{(S)} - T_{\text{TopKAttention}}^{(S)} \approx w (T_{\text{proj-}V} - T_{\text{proj-}V’}) + (S - w)(T_{\text{threshold}} - T_{\text{top-}k}) + T_{\text{power-law-fit}}
\end{equation}
Given $w \ll S$, the first term becomes negligible and the power-law fit is a
one-time operation using fast tensor operations. Thus, the runtime difference is
dominated by the last two terms:
\begin{equation}
T_{\text{\method}}^{(S)} - T_{\text{TopKAttention}}^{(S)} \approx (S - w) T_{\text{threshold}} - T_{\text{top-}k}
\end{equation}
Since thresholding can be done element-wise, in parallel and does not require
sorting like top-$k$, we have $T_{\text{top-}k} > T_{\text{threshold}}$, leading
a negative RHS in the above equation.

\subsection{Implementation Details}
\label{subsec:implementation-details}
Next, we discuss key implementation details of \method. We develop a
Triton-based kernel for the approximate generation phase. For the warmup phase
and power-law fit, we use standard PyTorch~\citep{paszke2017automatic} operations
with $\texttt{torch.compile}$ ~\citep{pytorch20} enabled for performance. A key
thing to note is that our triton kernel is not the most efficient implementation
of this approach. The reason being that, for an efficient implementation of
\method (and any top-$k$ based approach), the pruned attention scores and value
vectors need to be compacted in memory (Line 6,
Algorithm~\ref{alg:threshold-approx}). If this is not done, the resulting
vector-matrix product will be a sparse-sparse product, which is not efficient on
modern GPU architectures when compared to a dense vector-matrix products. Triton
does not allow for fine-grained manipulation of shared memory, leading us to
write the intermediate indices in global memory. This limitation is not present
if implemented directly in CUDA, and we leave this as future work. Due to this,
we will see later that our approach cannot compete with highly optimized fused
standard attention kernels. However, our implementation works as a
proof-of-concept, showing that our approach can lead to a reduction in HBM to
SRAM data movement, as seen in our experiments.

\section{Experimental Setup}
\label{sec:setup}
For model quality benchmarking, we evaluate \method on two categories of tasks:
perplexity and downstream generation. Evaluations are conducted on a subset of
open-weight models: Llama-3.1 (8B Instruct and 70B Instruct), Llama-3.2 3B
Instruct, Mistral 7B, Qwen2.5 7B~\citep{grattafiori2024llama3herdmodels,
jiang2023mistral, yang2024qwen2}. Due to tokenizer incompatibility, the Qwen
model is excluded from downstream generation tasks. The main text presents
results for the Llama family, while results for other models are provided in
appendix \ref{sec:appendix-results}.

We evaluate perplexity using the WikiText-2~\citep{wikitext-103} dataset.
Downstream generation evaluations are divided into short-context and
long-context settings. Short-context tasks include
IFEval~\citep{zhou2023instruction} and MATH-Hard ~\citep{fan2024hardmath},
selected from the Hugging Face Open LLM
Leaderboard~\citep{open-llm-leaderboard-v2}. The long-context setting employs
LongGenBench~\citep{wu2024longgenbench}, capped at 4096 tokens due to
computational constraints.

We compare against three baselines: Full Attention (no approximations), Top-$k$
Attention~\citep{topkattention}, and H$_2$O~\citep{zhang2023hH2O}. H$_2$O is
excluded for Llama-3.1-70B-Instruct due to its lack of multi-GPU support, and
from downstream generation tasks due to computational constraints. Top-$k$ is
evaluated only on perplexity and short-context tasks, due to computational
constraints. To ensure fair evaluation, we discard samples where the number of
generated tokens is less than the maximum warmup length in our experiments and
report metrics only on valid completions, for both \method and all baselines.
For Top-$k$ we use $k \in \{0.5,0.25,0.125, 0.05\}$, and an analogous retain
budget for H$_2$O. For \method, we use $\tau \in \{0.5,0.75,0.875,0.95\}$ and
warmup $w \in \{16,32,64,128,256,512\}$.

For runtime benchmarking, we evaluate our Triton-based implementation of the
approximate generation phase of \method  on a micro-benchmark, written using
Triton's benchmarking utilities. We compare our method against the fused scaled
dot-product attention kernel (SDPA) from PyTorch. We enable
\texttt{torch.compile} with \texttt{max-autotune} for both implementations. We
profile the memory transfer volume using Nsight Compute~\citep{ncu}. Model
quality runs are performed on GH200, and H100 GPUs. Runtime experiments are
performed on A100 GPUs. Bigger models are parallelized across 4 H100 GPUs, using
AxoNN~\citep{singh:sc2024, singh:ipdps2022}.

\section{Results}
\label{sec:results}
We now compare \method against full attention and other baselines, and evaluate
the performance of our Triton-based implementation relative to PyTorch’s fused
scaled dot-product attention kernels.

\begin{figure}[h]
  \centering
  \includegraphics[height=1.3in]{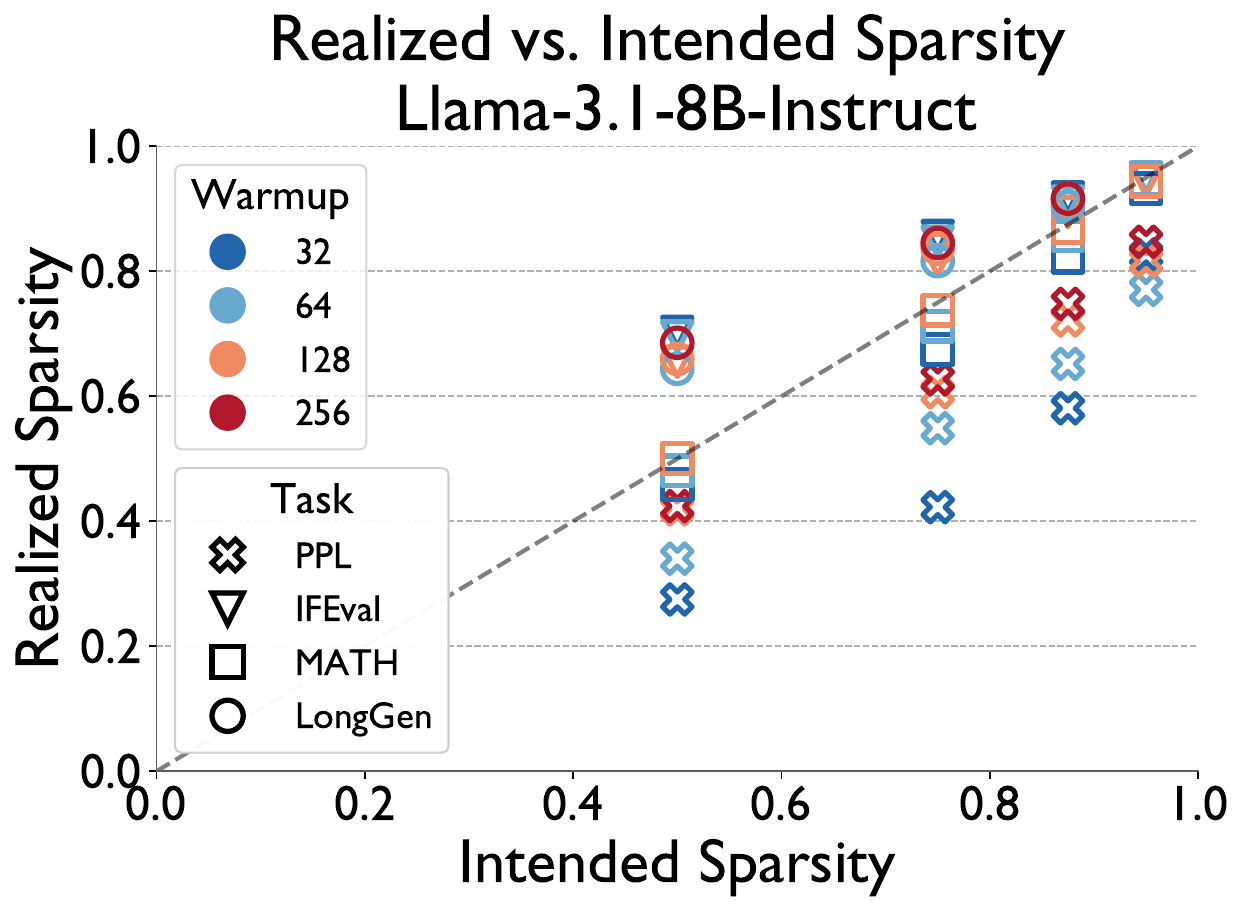}
  \includegraphics[height=1.3in]{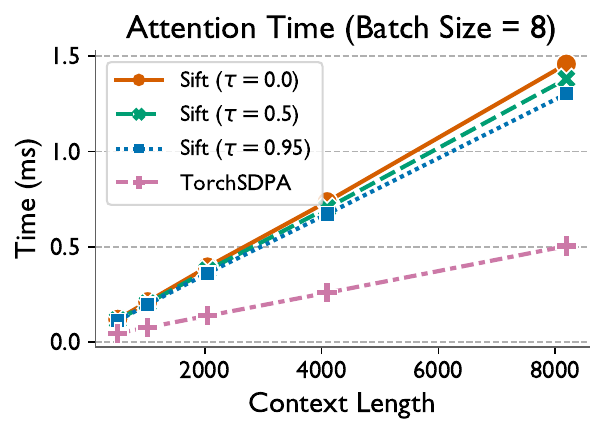}
  \includegraphics[height=1.3in]{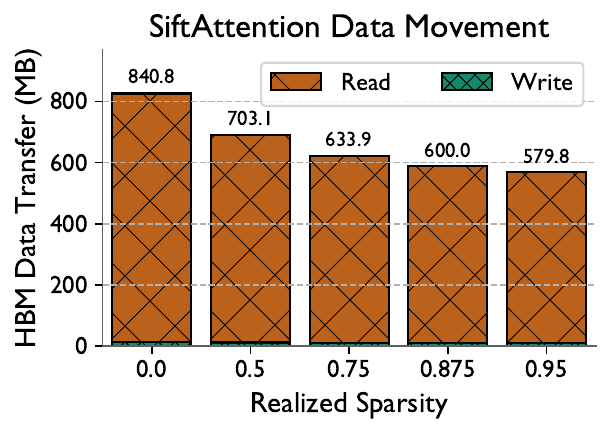}
  \caption{ (Left) \method Intended vs. Realized sparsity across tasks and
  warmup lengths for Llama-3.1–8B-Instruct. (Middle) Post-warmup attention
  latency of our Triton-based \method implementation, compared to PyTorch's fused
  SDPA across varying context lengths (Batch Size = 8). (Right) HBM data
  transfer volume as a function of realized sparsity (Context Length = 8192),
  showing reduced memory reads with increased sparsity.}
  \label{fig:sparsity}
\end{figure}

\textbf{Intended vs. Realized Sparsity:} We begin by distinguishing between
intended sparsity, set by hyper-parameters (e.g., $k$ in Top-$k$, $\tau$ in
\method), and realized sparsity, the actual fraction of filtered tokens during
inference. For static methods like Top-$k$ and H$_2$O, intended and realized
sparsity are equal by construction. However, in \method, sparsity is
determined dynamically based on thresholding learned from power-law fits and may
deviate from the intended sparsity.

We compute realized sparsity as the average ratio of pruned keys to total keys
across all generation steps and samples. Figure~\ref{fig:sparsity} (left) shows
this relationship for \method on Llama-3.1-8B-Instruct across various tasks and
warmup lengths. Perplexity-oriented tasks tend to undershoot the intended
sparsity, while downstream tasks like IFEval and LongGenBench often surpass it.
While we do not yet have a theoretical explanation for these task-specific
trends, a consistent observation is that longer warmup phases yield realized
sparsity closer to the target. This aligns with earlier findings that longer
warmups lead to better power-law fits for attention quantiles. Going forward, we
report realized sparsity when comparing \method with baselines to keep
comparisons fair. 

\textbf{Runtime and Data Transfer Volume Evaluation:} Figure~\ref{fig:sparsity}
(middle) compares \method's runtime (post-warmup) against PyTorch's fused Scaled
Dot-Product Attention (SDPA) kernel (exact attention). As discussed in
Section~\ref{subsec:implementation-details}, due to Triton's limited support for
shared memory and compaction, our kernel is slower. However, \method's runtime
improves with sparsity: from 1.46 ms to 1.30 ms (10\% reduction) at context
length 8192. Investigating further, Figure~\ref{fig:sparsity} (right) shows that
this gain stems from reduced HBM data transfer—falling from 840.8 MB to 579.8 MB
(31\% reduction) with increasing sparsity. This demonstrates that \method is
able to reduce the HBM to SRAM data movement, as well as other top-$k$ based
approaches. But while our Triton implementation does not fully exploit this, a
CUDA backend could.

\begin{figure}[ht]
  \centering
  \includegraphics[width=0.32\textwidth]{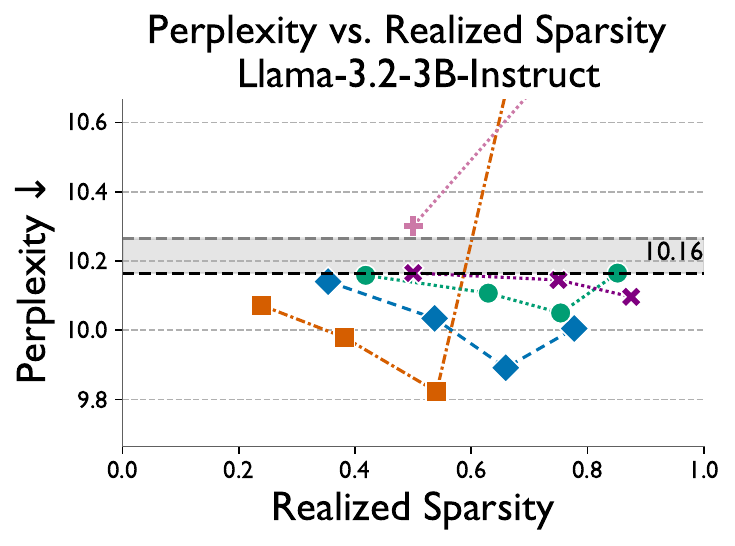}
  \includegraphics[width=0.32\textwidth]{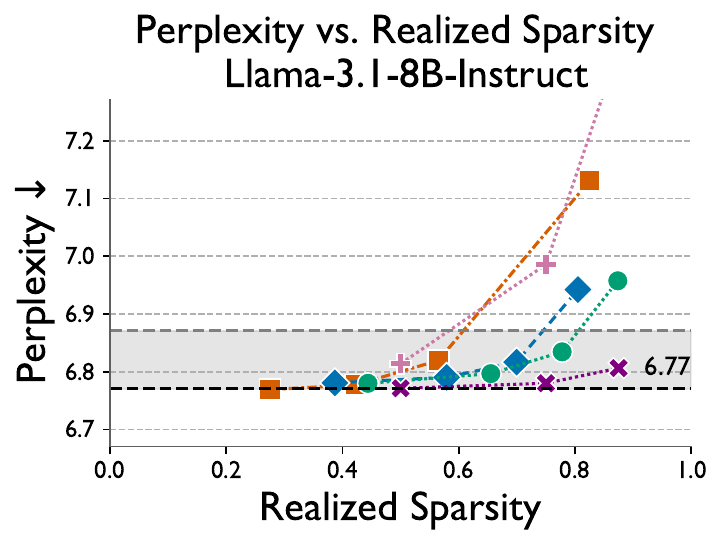}
  \includegraphics[width=0.32\textwidth]{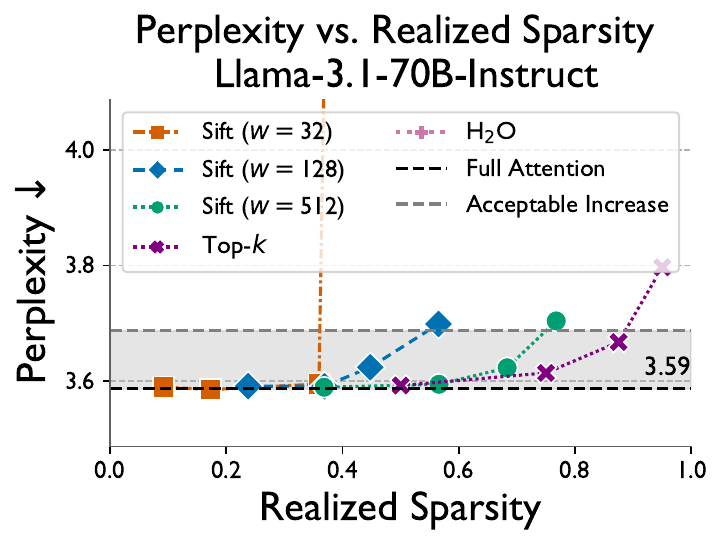}
  \caption{Perplexity evaluation (WikiText-2) comparing \method with baselines for 3 models}
  \label{fig:ppl_evaluation}
\end{figure}

\textbf{Perplexity Evaluation:} We now turn to Figure~\ref{fig:ppl_evaluation},
which compares perplexity on WikiText-2 for \method, full attention, and other
baselines across three Llama models. Focusing on Llama-3.1-8B-Instruct (center),
we see that longer warmup phases yield better perplexity for \method, due to
improved power-law fits. A short warmup of 32 steps (orange curve) leads to a
noticeable degradation, while warmups of 128 and 512 (green, blue) keep
degradation within 0.1~\citep{Yao2023ZeroQuantV2EP} for most sparsity levels,
rising to 0.2 only at the highest sparsity. H$_2$O performs worst, with drops
exceeding 0.5. Top-$k$ maintains consistently strong performance, even at the
highest realized sparsity values. However, with a sufficient warmup, \method
approaches Top-$k$ performance closely. Similar trends are observed for
Llama-3.1-70B-Instruct (right). Interestingly, for Llama-3.2-3B-Instruct (left),
both \method and Top-$k$ achieve lower perplexity than the full attention
baseline, with \method outperforming Top-$k$. Across all models, \method remains
in the acceptable range of perplexity degradation for most sparsity levels.
Results for other models are similar and can be found in appendix \ref{sec:appendix-results}.

\begin{figure}[ht]
  \centering
  \includegraphics[width=0.32\textwidth]{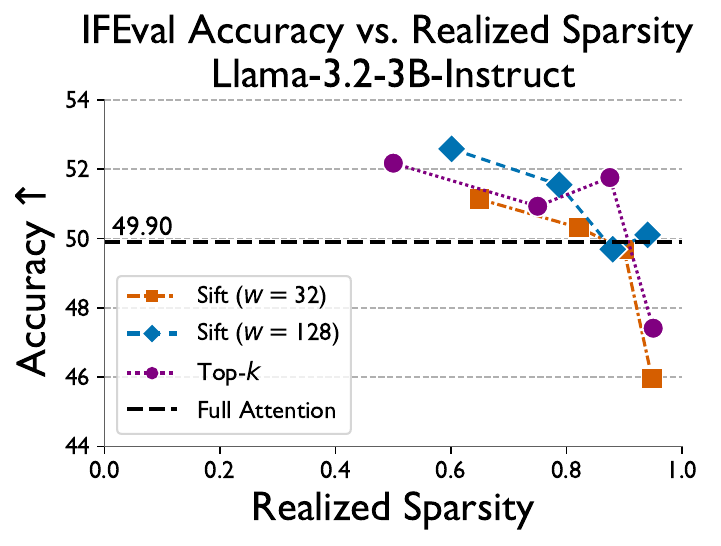}
  \includegraphics[width=0.32\textwidth]{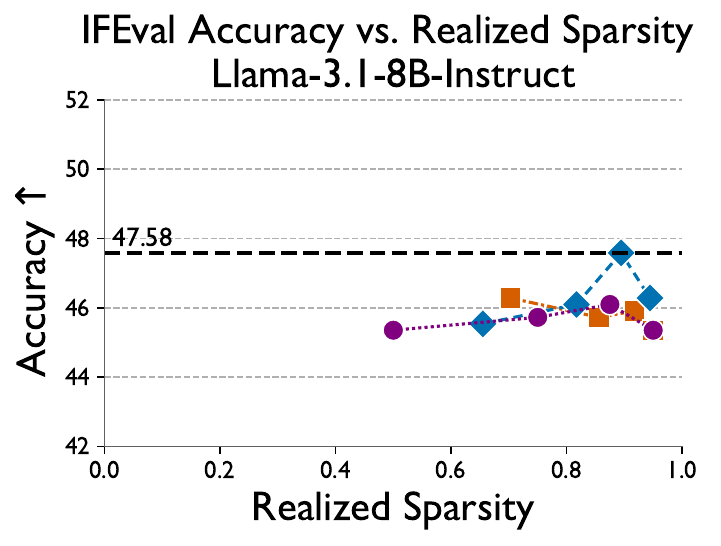}
  \includegraphics[width=0.32\textwidth]{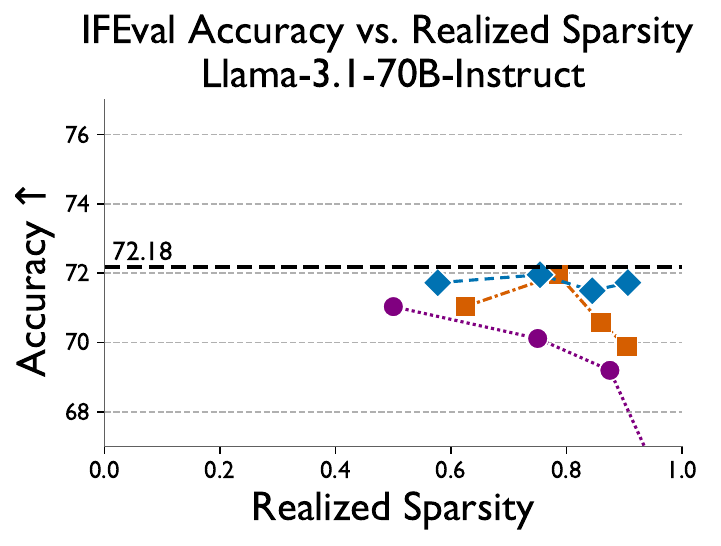}
  \includegraphics[width=0.32\textwidth]{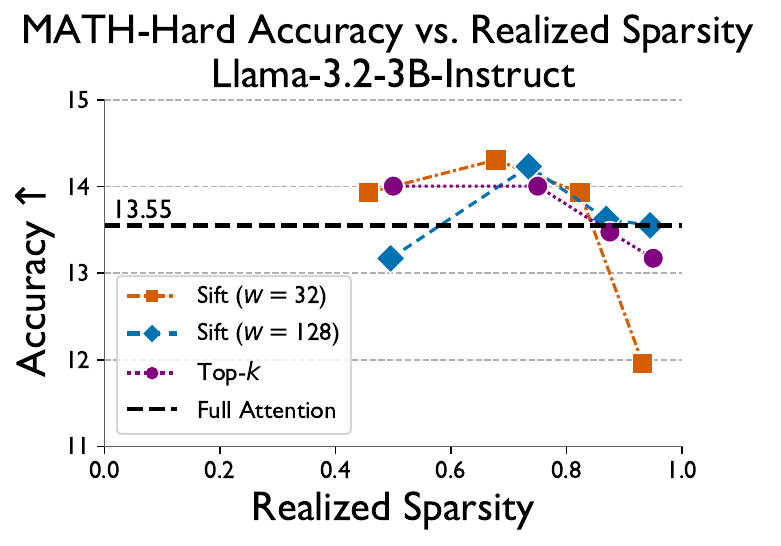}
  \includegraphics[width=0.32\textwidth]{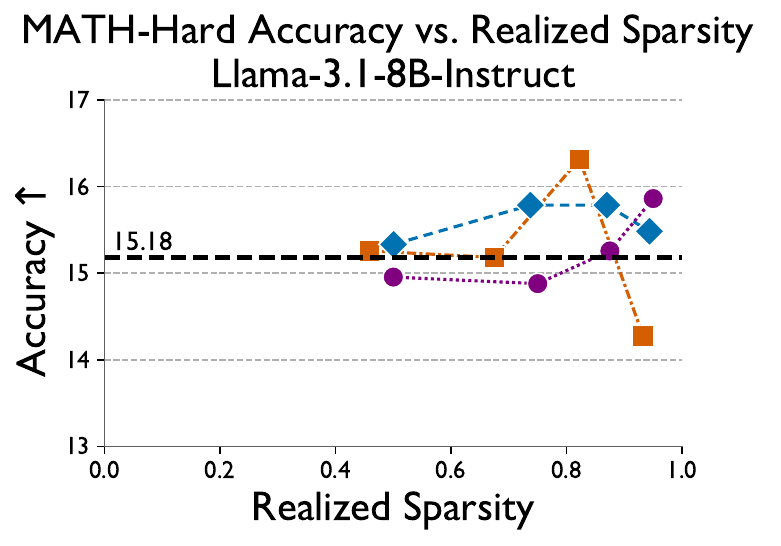}
  \includegraphics[width=0.32\textwidth]{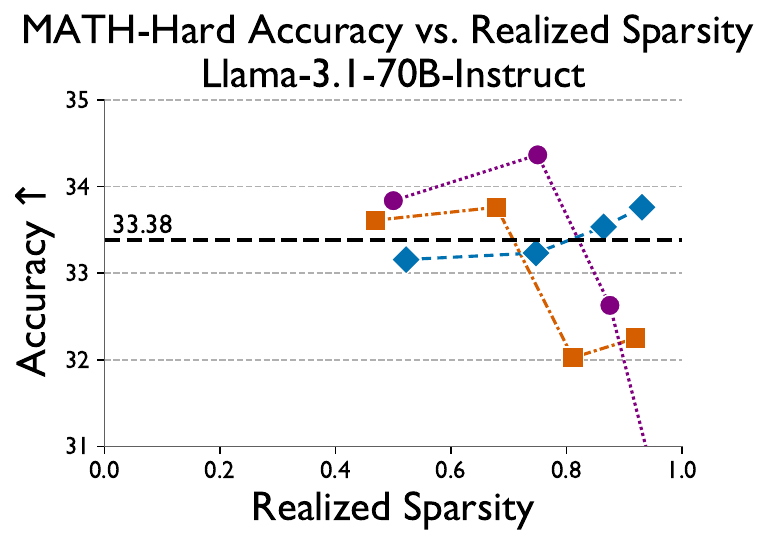}
  \includegraphics[width=0.32\textwidth]{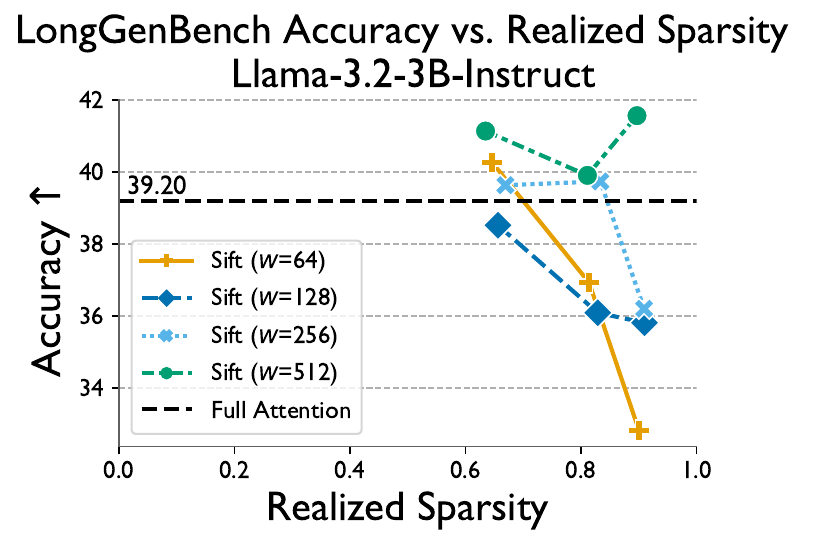}
  \includegraphics[width=0.32\textwidth]{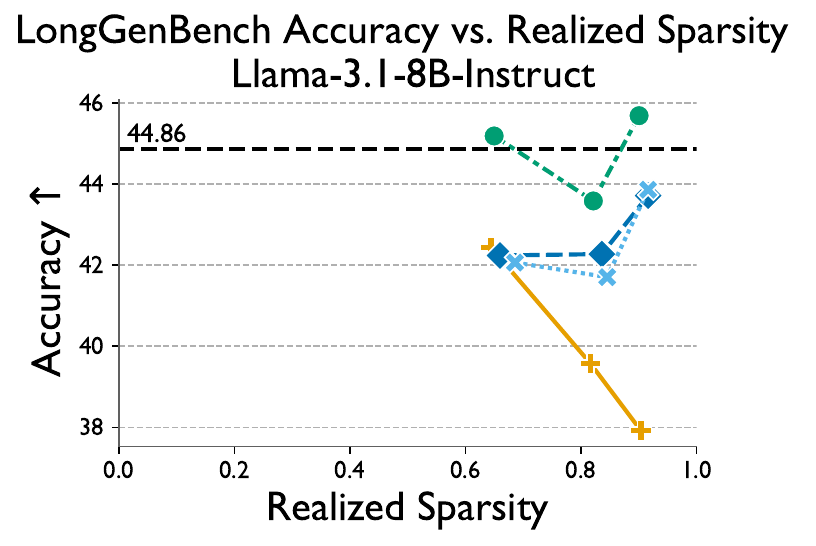}
  \includegraphics[width=0.32\textwidth]{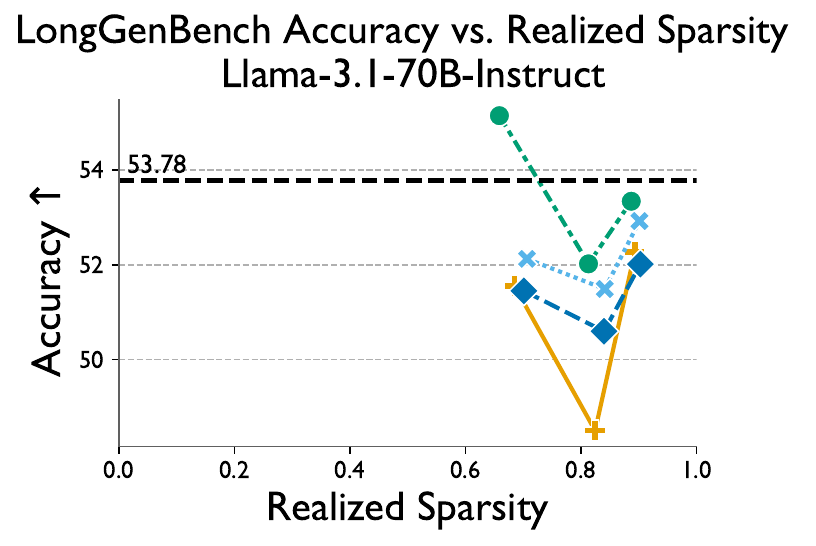}
  \caption{Short (IFEval, MATH) and long (LongGenBench) generation evaluation across 3 models}
  \label{fig:short_long_context_eval}
\end{figure}

\textbf{Downstream Generative Task Evaluation:} Finally, we evaluate \method on
downstream generative tasks—IFEval, MATH-Hard, and LongGenBench—shown in
Figure~\ref{fig:short_long_context_eval}. IFEval and MATH-Hard involve shorter
generation lengths (up to 1280 and 1024 tokens), while LongGenBench requires
generating up to 4096 tokens. For IFEval (top row), focusing on
Llama-3.2-3B-Instruct (left), we see that \method outperforms the Top-$k$
baseline given a sufficient warmup of 128 steps (blue curve), especially at high
realized sparsities. This trend is consistent across the other models. On
MATH-Hard (middle row), a similar trend can be observed, \method closely
matching or outperforming the Top-$k$ baseline. particularly, for
Llama-3.1-70B-Instruct, we see a severe degradation in performance with Top-$k$
at high sparsities, but \method maintains strong performance. Finally, on
LongGenBench (bottom row), \method maintains strong performance even with the
longer generation lengths. The average drop in accuracy is 2-3\% with a warmup
of 256 steps (teal blue curve) across models at high sparsity values. With 512
warmup steps, it nearly matches the full attention baseline.

Together with the perplexity results, these findings highlight \method's ability
to retain strong task performance across models and sparsity levels, often
outperforming Top-$k$, while being easier to parallelize on GPUs.


\section{Conclusion}
\label{sec:conclusion}
This work introduces \method, a dynamic power-law guided sparse attention algorithm 
that reduces GPU memory movement between HBM and SRAM.  We observe that attention 
score quantiles follow a predictable power-law decay over generation steps, consistently 
across models and datasets - a novel finding that can motivate future work in
sparse attention methods. Leveraging this insight, \method uses a brief warmup
phase to fit a power-law model, then predicts thresholds to prune attention
scores without requiring costly top-$k$ sorting. Our experiments show that
\method maintains model quality, similar to or better than existing sparse
attention techniques, while reducing the data movement between the GPU HBM and
SRAM.

\subsection{Limitations and Future Work} 
Our Triton-based implementation of \method is not fully optimized and performs
sub-optimally compared to PyTorch's SDPA kernel. This shortfall arises from inherent 
limitations in Triton and inefficiencies in sparse vector-matrix multiplication.
To address some of these performance issues, we are actively developing a CUDA-based
kernel. Additionally, \method relies on a fixed-length warmup phase,
necessitating careful tuning specific to each model and task. An interesting
future direction is the exploration of more advanced online fitting techniques,
allowing the power-law parameters to be adaptively re-calibrated throughout
generation based on discrepancies between predicted and observed attention
quantiles. Finally, while our current power-law model is based on post-softmax
attention scores, future work could focus on estimating thresholds directly from
pre-softmax scores, potentially yielding further computational efficiencies.

\bibliographystyle{plain}
\bibliography{./bib/cite,./bib/pssg}

\clearpage
\appendix
\section{Extended Power-Law Analysis Results}
\label{sec:extended-power-law-analysis}
\subsection{Power-Law Samples}
\begin{figure}[hbt!]
  \centering
  \includegraphics[width=0.245\linewidth]{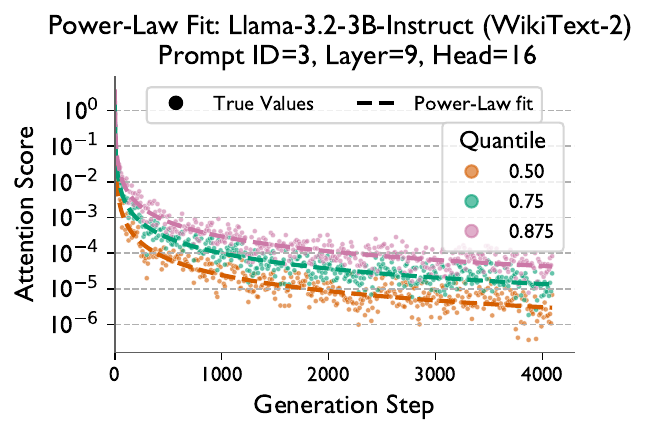}
  \includegraphics[width=0.245\linewidth]{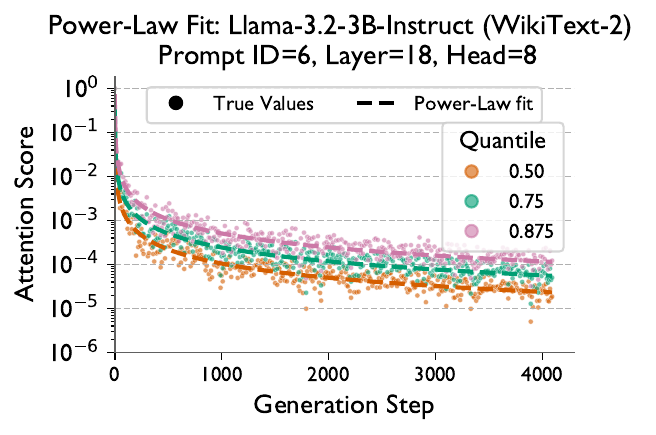}
  \includegraphics[width=0.245\linewidth]{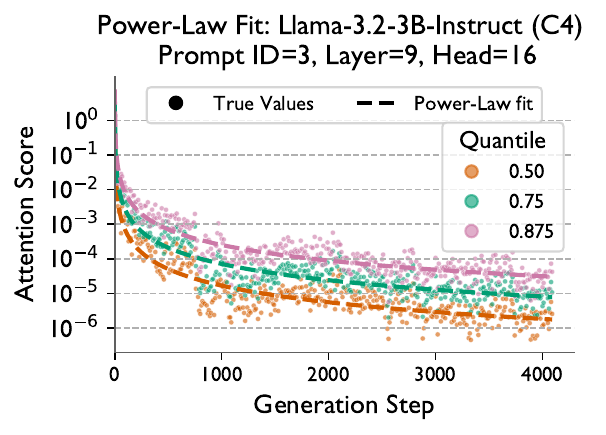}
  \includegraphics[width=0.245\linewidth]{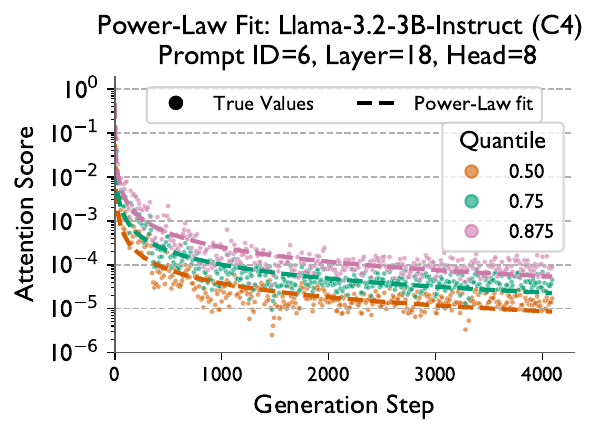}
  \includegraphics[width=0.245\linewidth]{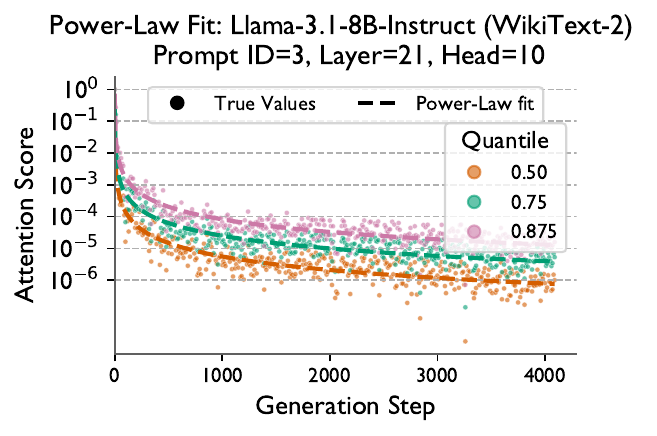}
  \includegraphics[width=0.245\linewidth]{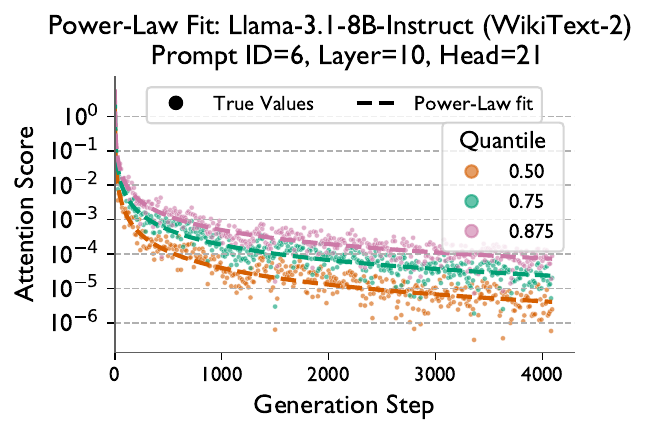}
  \includegraphics[width=0.245\linewidth]{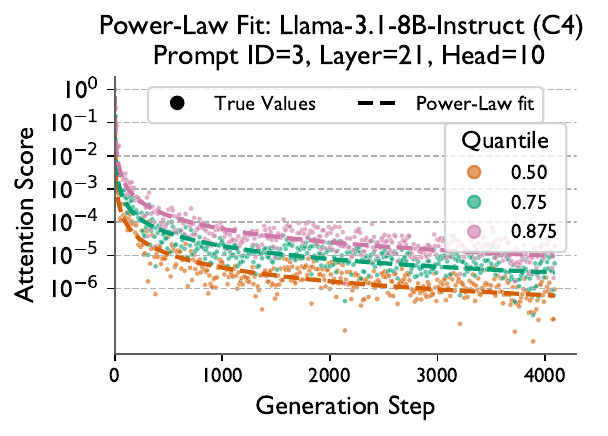}
  \includegraphics[width=0.245\linewidth]{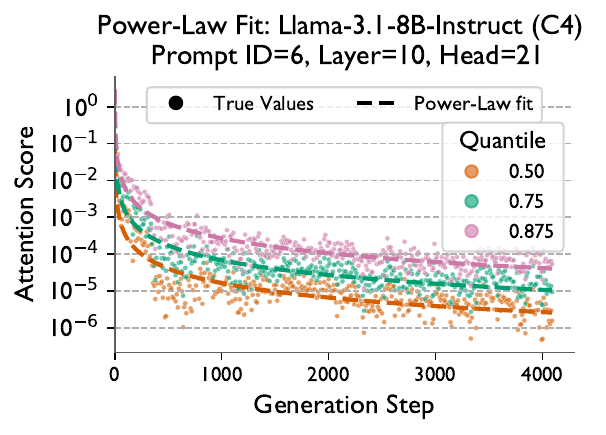}
  \includegraphics[width=0.245\linewidth]{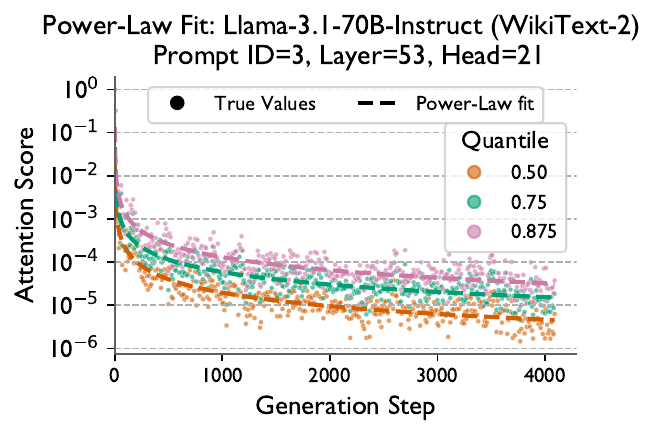}
  \includegraphics[width=0.245\linewidth]{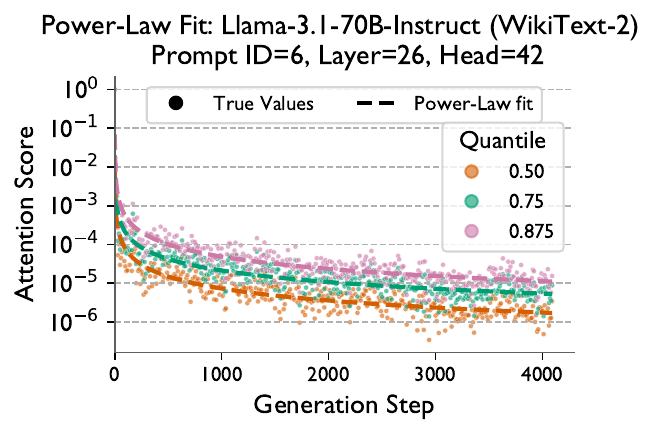}
  \includegraphics[width=0.245\linewidth]{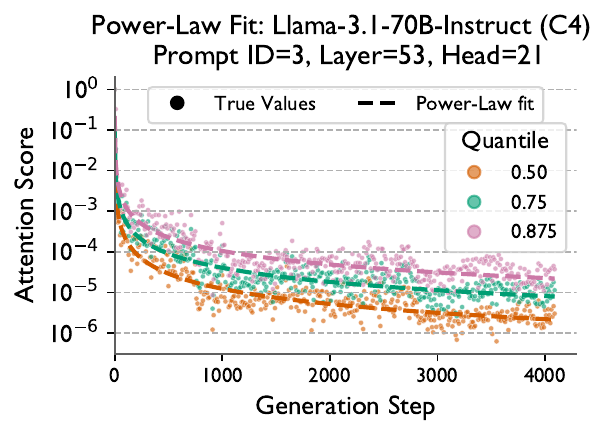}
  \includegraphics[width=0.245\linewidth]{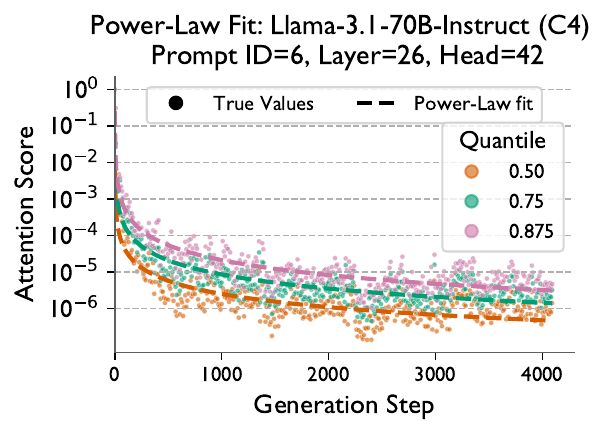}
  \includegraphics[width=0.245\linewidth]{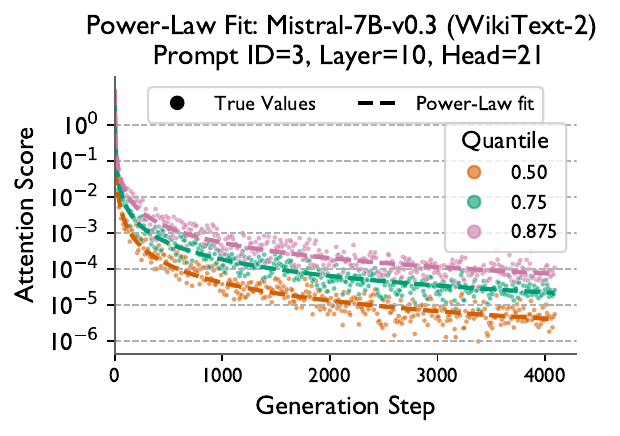}
  \includegraphics[width=0.245\linewidth]{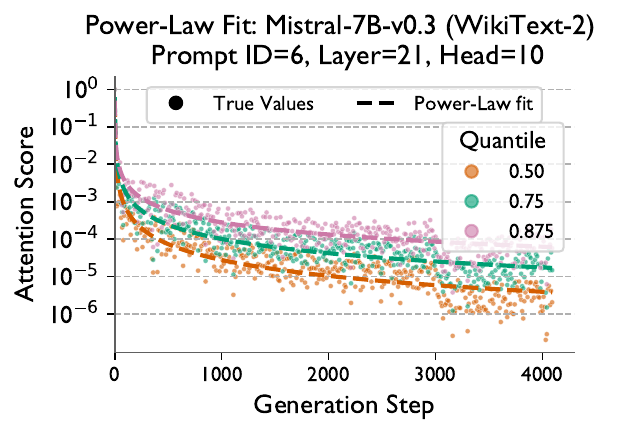}
  \includegraphics[width=0.245\linewidth]{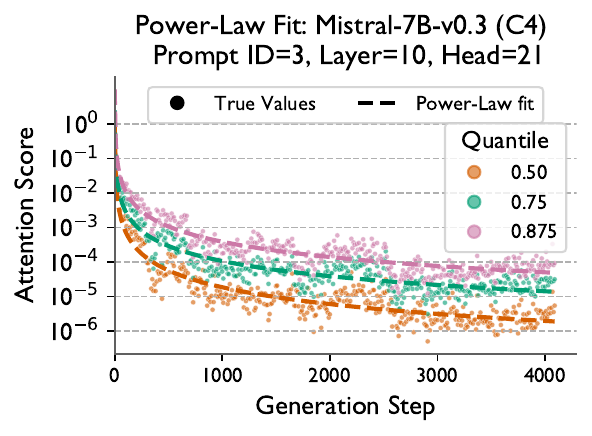}
  \includegraphics[width=0.245\linewidth]{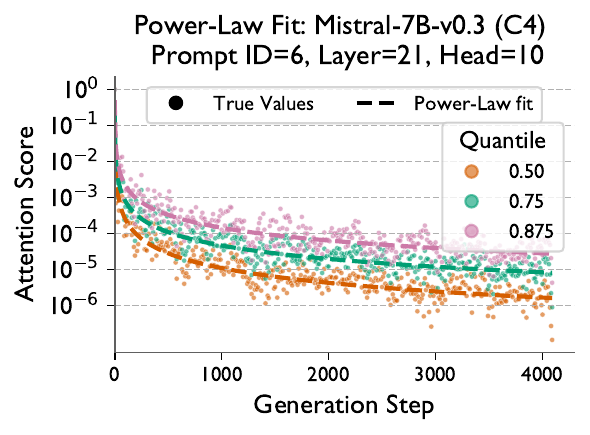}
  \includegraphics[width=0.245\linewidth]{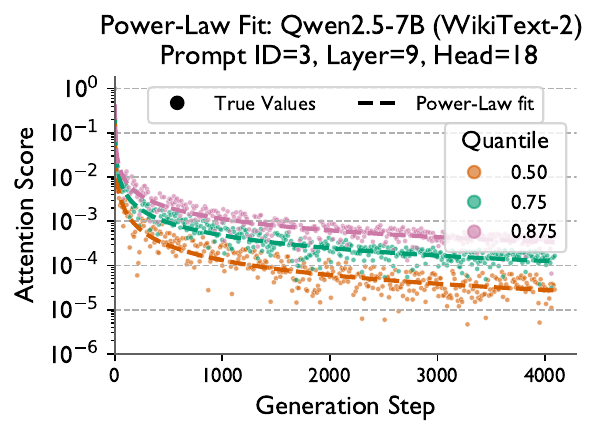}
  \includegraphics[width=0.245\linewidth]{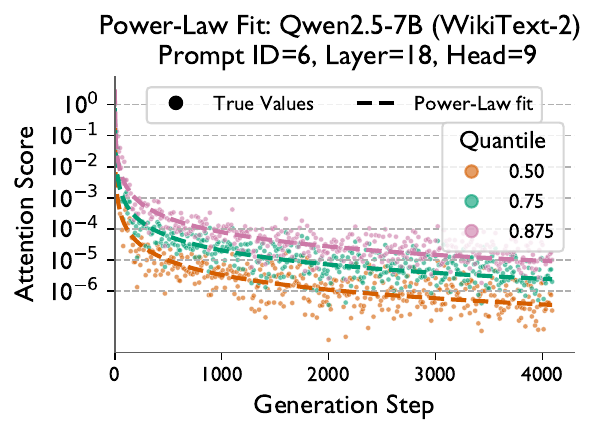}
  \includegraphics[width=0.245\linewidth]{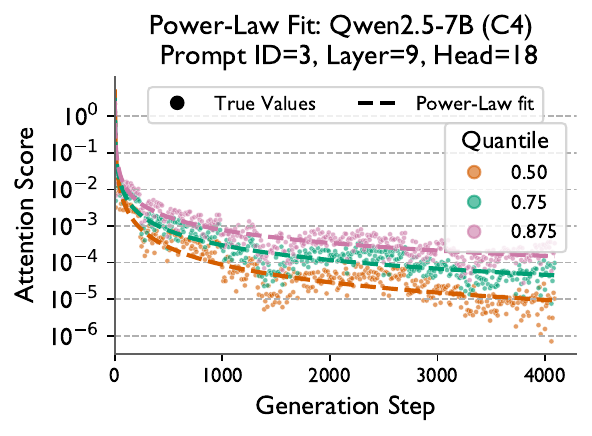}
  \includegraphics[width=0.245\linewidth]{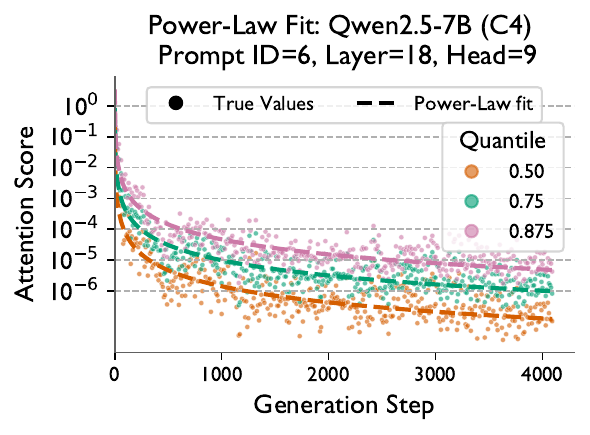}
  \caption{Power-Law fit for attention score quantiles (log y-scale) over
  generation steps across 5 different models. The two left columns are samples
  on WikiText-2, and the two right columns are samples on C4.}
  \label{fig:appendix-powerlaw-fit-sample}
\end{figure}
In this section, we present extended results from our power-law analysis of
attention score quantiles. Figure~\ref{fig:appendix-powerlaw-fit-sample}
displays representative samples of the observed quantiles across various models,
datasets, prompts, layers, and attention heads. These qualitative results
suggest that a power-law trend can often be visually inferred from the evolution
of attention score quantiles. While the trend appears weaker in some samples
from the C4 dataset, this may be attributable to suboptimal sample selection
rather than a lack of underlying structure.

Although this analysis is not exhaustive, it offers evidence that power-law
behavior is a consistent characteristic of attention score distributions. As
reported in Figure~\ref{fig:powerlaw-fit} of the main text, the $R^2$
values—indicating the goodness-of-fit for the power-law model—consistently
exceed 0.6 across all evaluated models and datasets. Notably, the C4 dataset
exhibits tighter variance in $R^2$ values.

\subsection{Power-Law Fitting with Warmup}
\begin{figure}[hbt!]
  \centering
  \includegraphics[width=0.35\linewidth]{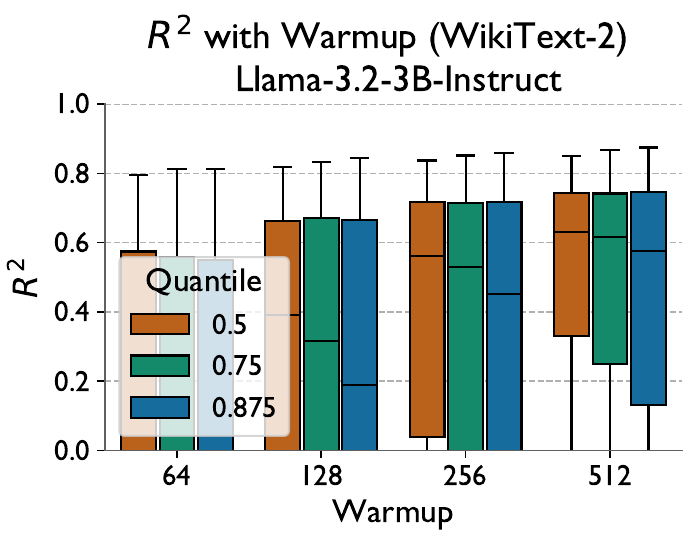}
  \includegraphics[width=0.35\linewidth]{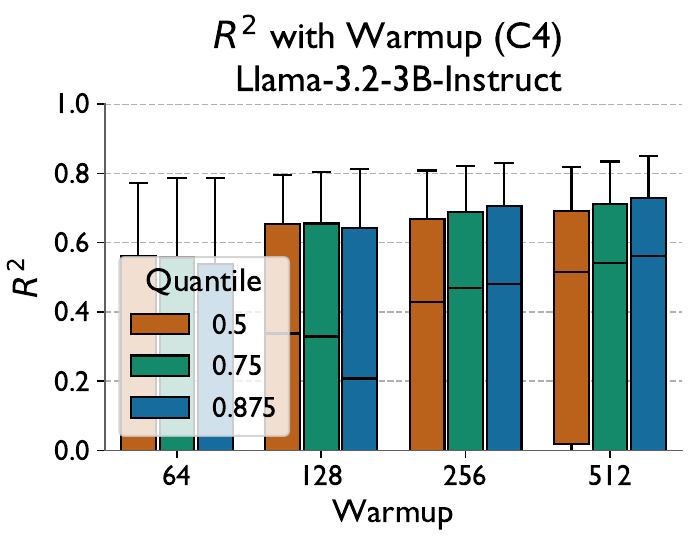}
  \includegraphics[width=0.35\linewidth]{figs/powerlaw-fit/wikitext-test/Llama-3.1-8B-Instruct_warmup_r2_box.pdf}
  \includegraphics[width=0.35\linewidth]{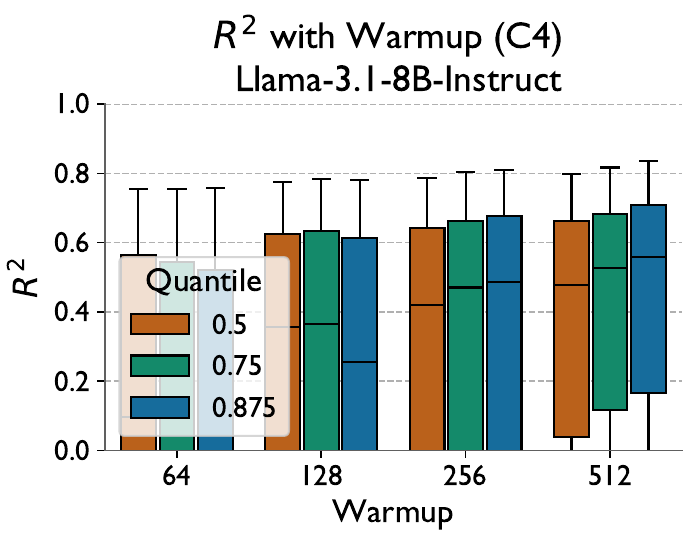}
  \includegraphics[width=0.35\linewidth]{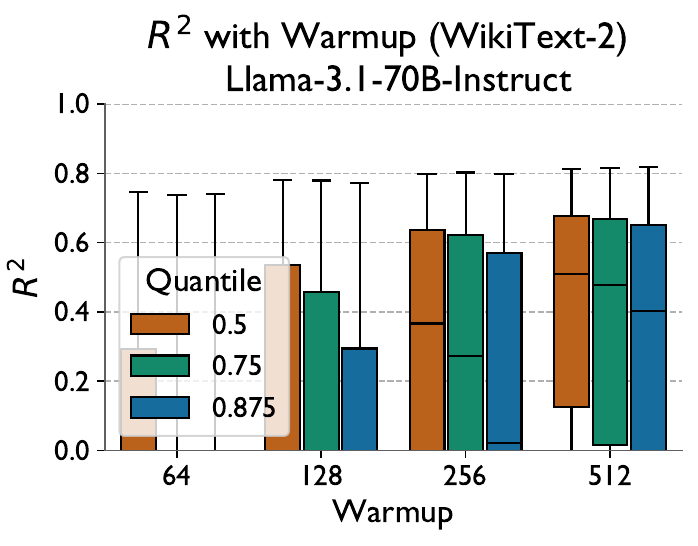}
  \includegraphics[width=0.35\linewidth]{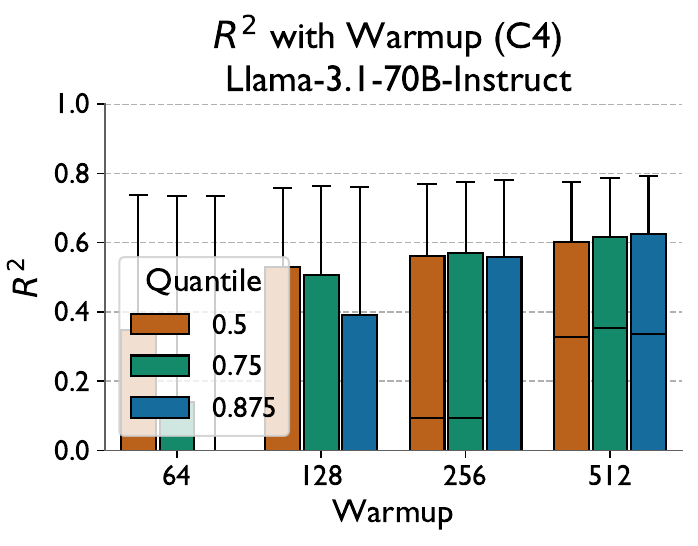}
  \includegraphics[width=0.35\linewidth]{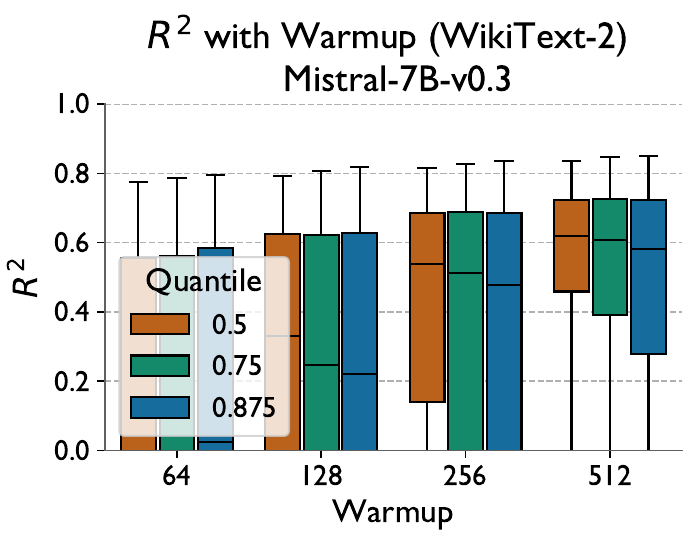}
  \includegraphics[width=0.35\linewidth]{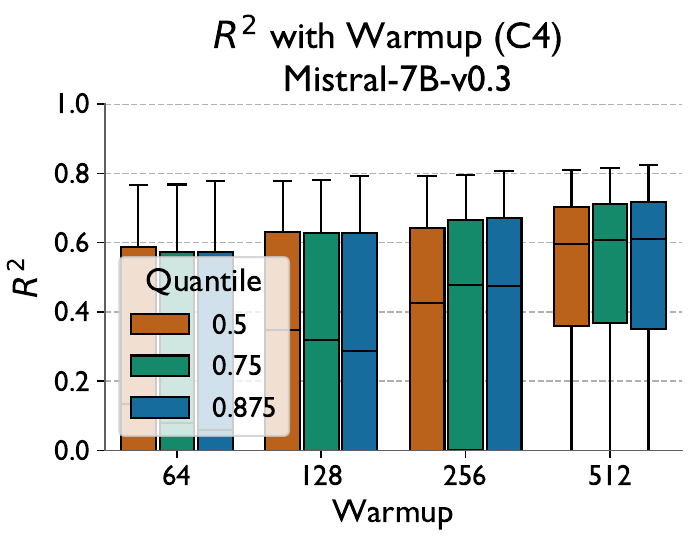}
  \includegraphics[width=0.35\linewidth]{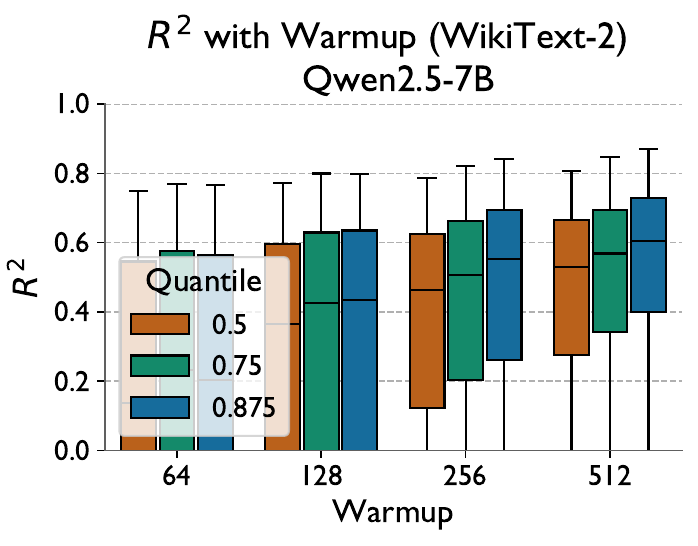}
  \includegraphics[width=0.35\linewidth]{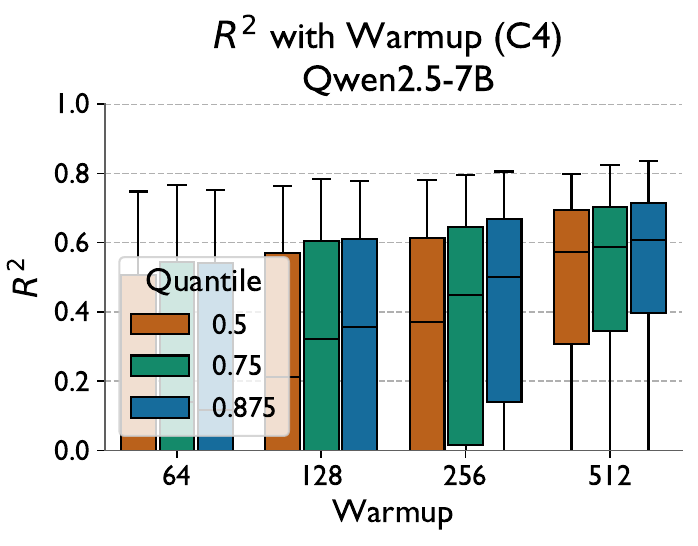}
  \caption{$R^2$ distribution across 5 different models,
      evaluated across warmup sizes $w$: 64, 128, 256, and 512 steps. Left
      columns are samples on WikiText-2; right columns are samples on C4}
  \label{fig:appendix-powerlaw-fit-r2-warmup}
\end{figure}
In this section, we further investigate how the length of the warmup period
affects the estimation of the power-law fit parameters, for a
perplexity evaluation task.

Figure~\ref{fig:appendix-powerlaw-fit-r2-warmup} shows a clear correlation
between warmup length and power-law fit quality. With shorter warmup periods
(e.g., 64 steps), we observe significantly lower $R^2$ scores—often dropping
into negative values. These negative values indicate that the power-law fit
performs worse than a naive model that simply predicts the mean of the
log-quantile values, especially highlighting unreliable parameter estimation
with short warmup periods. This degradation is especially evident in the C4
dataset, where initial $R^2$ scores are consistently lower than those observed
on WikiText-2.

As the warmup length increases, the fit quality steadily improves across all
datasets. By 512 warmup steps, $R^2$ values become consistently positive and
substantially higher, indicating that a longer warmup allows for more accurate
modeling of the attention score quantile decay. This trend underscores the
importance of allocating a sufficient warmup period for reliable threshold
estimation in \method. We also observe that the fit quality is better on
WikiText-2 than on C4. While these results demonstrate the feasibility of 
the warmup approach in \method, we acknowledge that the fitting accuracy could
be improved. More sophisticated power-law fitting algorithms could be 
explored in future work, to bridge the gap in estimating the real power-law trend
with a high $R^2$ (as shown in the previous section). One possible approach 
could explore using realized sparsity signals to implement an adaptive warmup
schedule

Despite occasional poor fit quality, we observe strong downstream performance
even with shorter warmup lengths, as shown in Section~\ref{sec:results} and
Appendix~\ref{sec:appendix-results}. We hypothesize that this robustness arises
because, although the realized sparsity may deviate from the intended target,
the model still retains a meaningful and consistent subset of top-$\tau$
attention scores. In effect, the power-law may be accurately modeling a
different effective sparsity level than initially specified.

Another limitation of this analysis is that we only analyze the power-law fit on
a perplexity task. We observe that the gap between realized and intended
sparsity is larger on perplexity evaluation than on other tasks such as IFEval
and MATH. We hypothesize that these results are linked and that the same
analysis performed on other tasks like IFEval and MATH could yield significantly
better $R^2$ scores.

\section{Detailed Evaluation Results}
\label{sec:appendix-results}
\subsection{Realized vs. Intended Sparsity}
\begin{figure}[hbt!]
  \centering
  \includegraphics[width=0.45\linewidth]{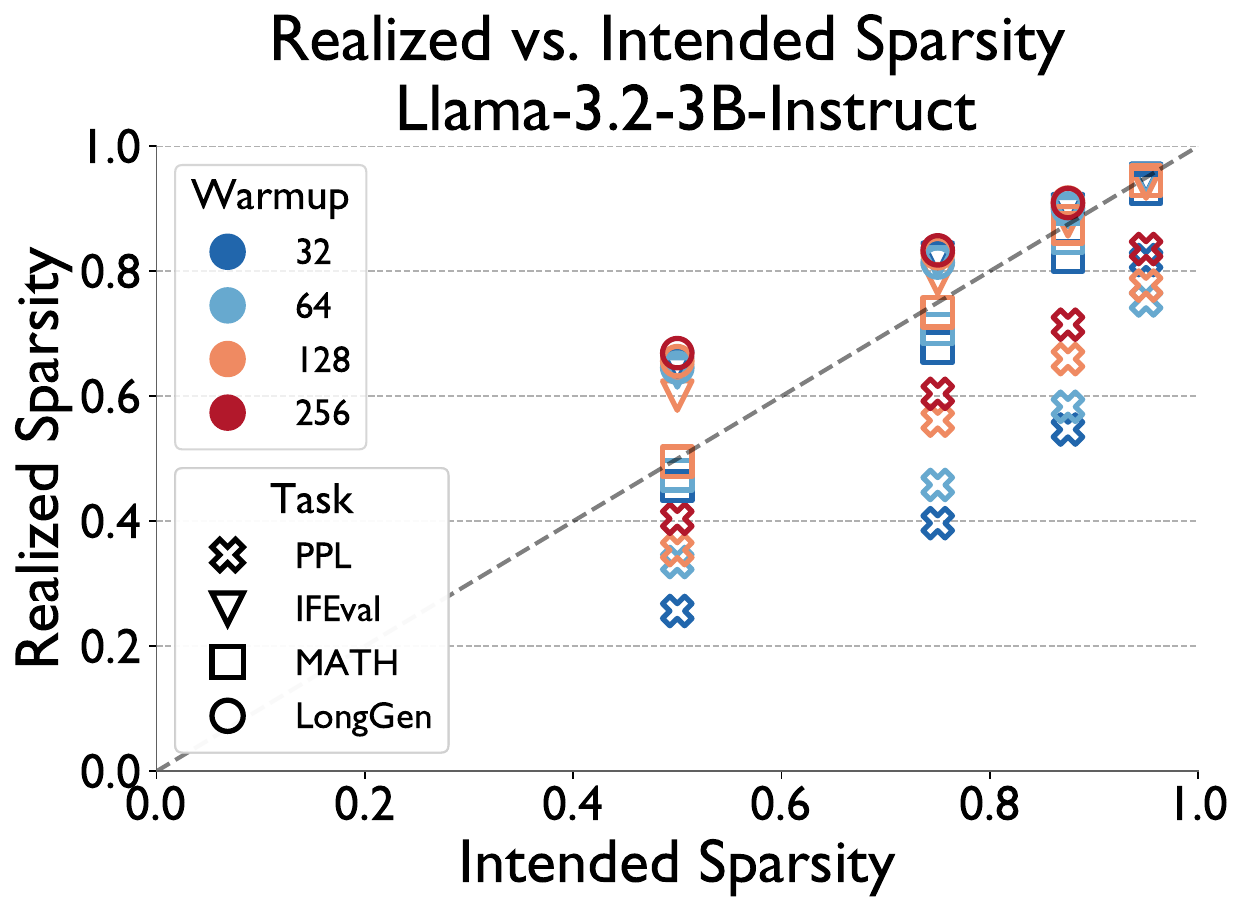}
  \includegraphics[width=0.45\linewidth]{figs/ml-evaluation/figures/main/main_realized_vs_intended_sparsity_Llama-3.1-8B-Instruct_thresh.pdf}
  \includegraphics[width=0.45\linewidth]{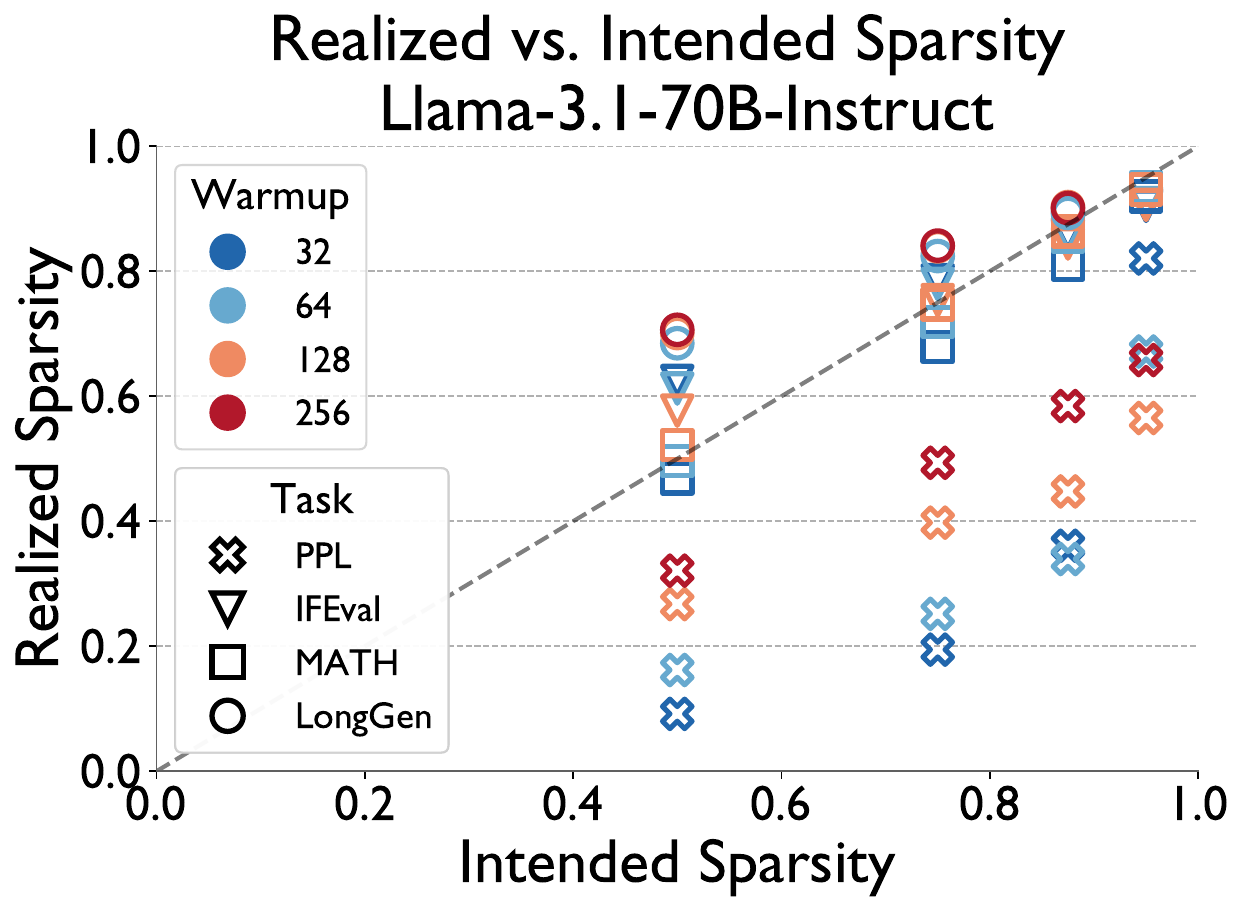}
  \includegraphics[width=0.45\linewidth]{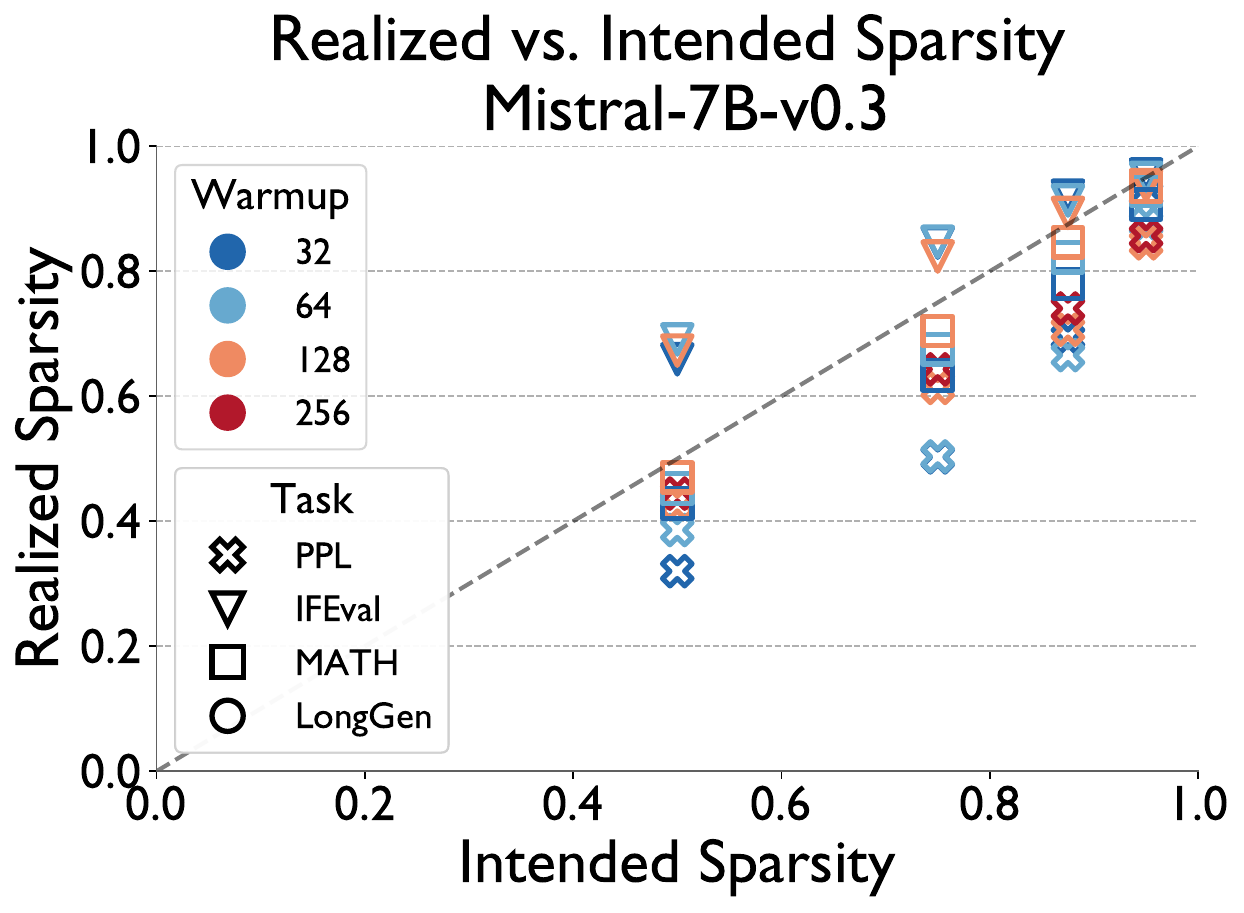}
  \caption{Intended vs. Realized sparsity on \method across tasks and
  warmup lengths for 5 different models.}
  \label{fig:appendix-realized-intended-sparsity}
\end{figure}
In this section we analyze the relationship between intended and realized
sparsity for \method across multiple models, tasks, and warmup lengths, as shown
in Figure~\ref{fig:appendix-realized-intended-sparsity}. We observe that the
disparity between intended and realized sparsity varies significantly across
models. Generation tasks consistently demonstrate realized sparsity closer to
the intended sparsity, suggesting a potential relationship between task
type and sparsity behavior. This trend warrants further investigation.

Notably, the Llama-3.1-70B model exhibits a weaker correlation between realized
and intended sparsity compared to smaller models. While this could indicate
different underlying dynamics in larger models, these results are not conclusive
and require additional study to fully understand the scaling behavior of
attention patterns.

The effectiveness of \method is evident across all models: increasing the warmup
length consistently brings realized sparsity closer to intended levels.
Furthermore, we observe that increasing the intended sparsity hyper-parameter
(specified by the $\tau$-th quantile) leads to proportional increases in
realized sparsity, demonstrating the method's ability to effectively control
sparsification across different scales and architectures.

\subsection{Perplexity Evaluation}
\begin{figure}[h]
  \centering
  \includegraphics[width=0.45\linewidth]{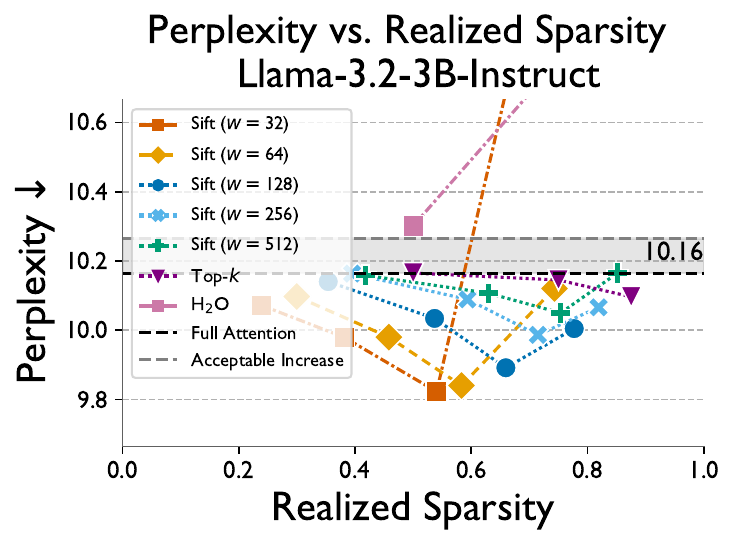}
  \includegraphics[width=0.45\linewidth]{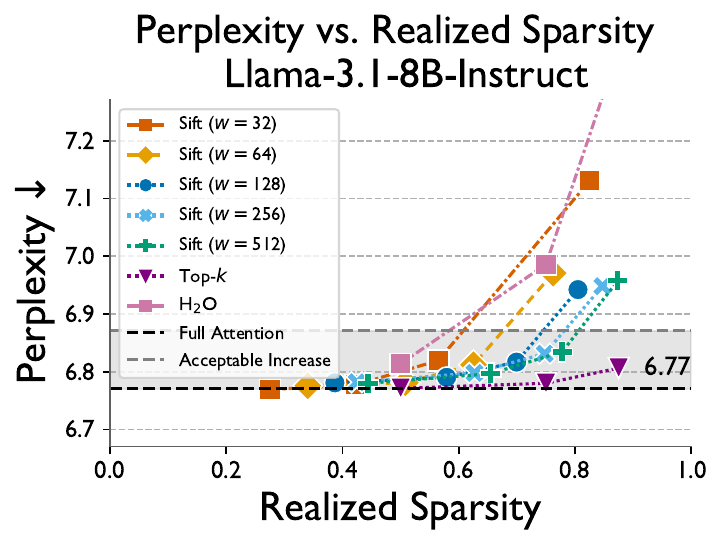}
  \includegraphics[width=0.45\linewidth]{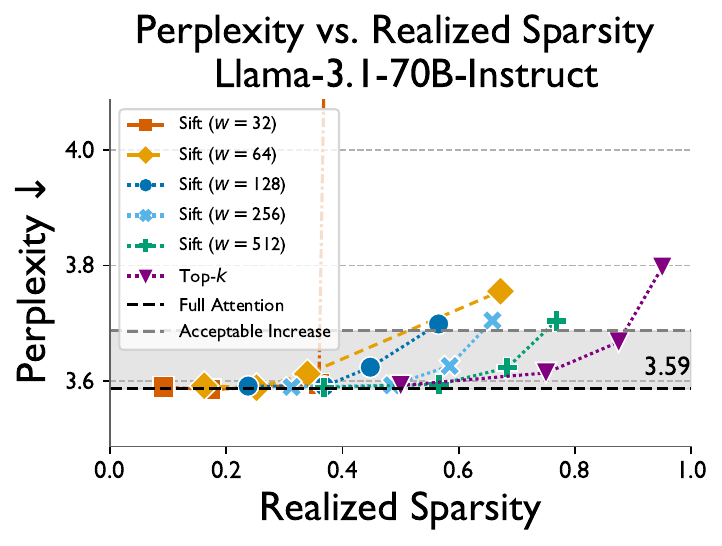}
  \includegraphics[width=0.45\linewidth]{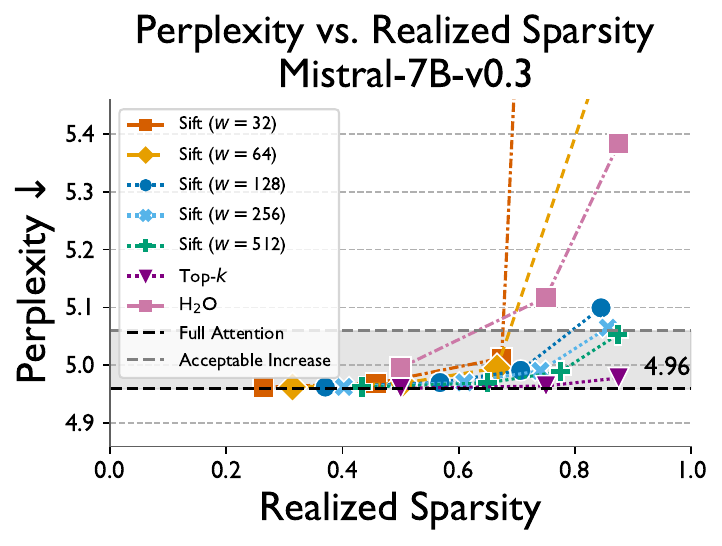}
\includegraphics[width=0.45\linewidth]{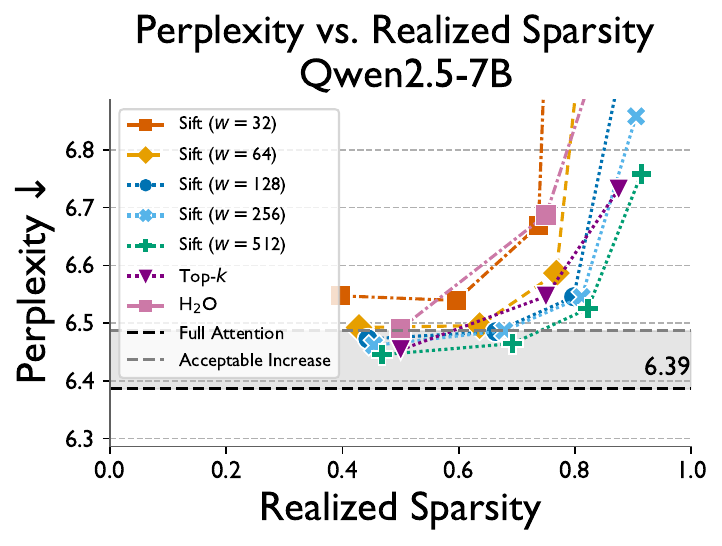}
  \caption{Perplexity evaluation (WikiText-2) comparing \method with baselines across 5 models}
  \label{fig:appendix-ppl}
\end{figure}

Figure~\ref{fig:appendix-ppl} presents the perplexity evaluation results on
WikiText-2 across five different models. Consistent with the results shown in
Figure~\ref{fig:ppl_evaluation} of the main text, we find that \method achieves
perplexity comparable to or better than the Top-$k$ baseline (dark purple). The
H$_2$O method (light pink) performs the worst, with the perplexity degradation
exceeding acceptable levels—even at low realized sparsity. Additionally, we
observe that across all models, longer warmup lengths correlate with improved
perplexity, underscoring the importance of sufficient warmup for accurate
threshold estimation.

\subsection{Short Context Tasks}
\begin{figure}[h]
  \centering
  \includegraphics[width=0.45\textwidth]{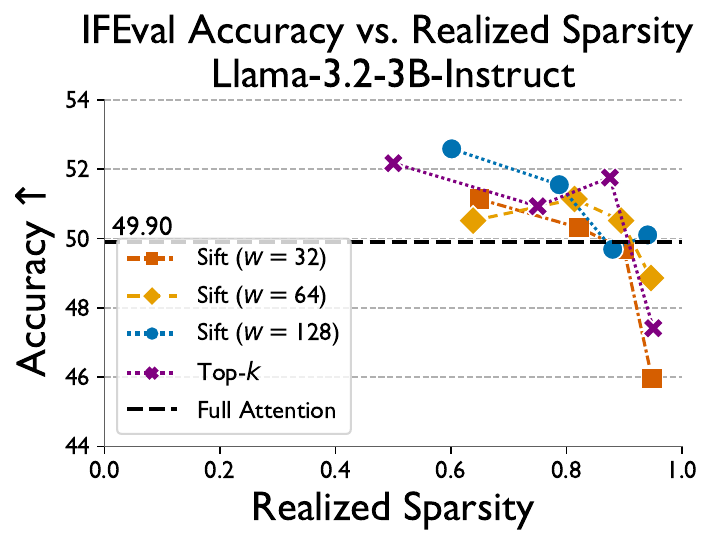}
  \includegraphics[width=0.45\textwidth]{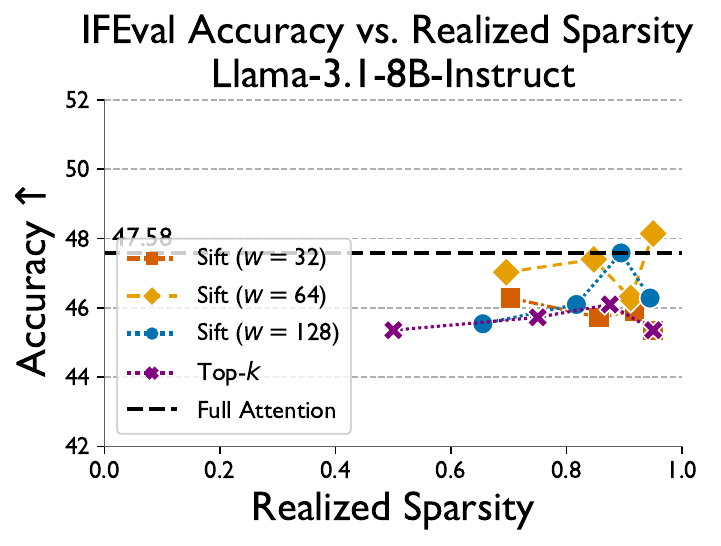}
  \includegraphics[width=0.45\textwidth]{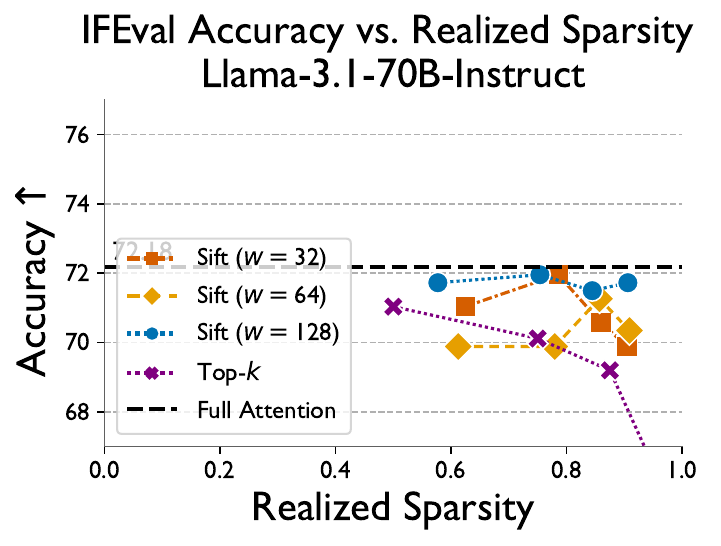}
  \includegraphics[width=0.45\textwidth]{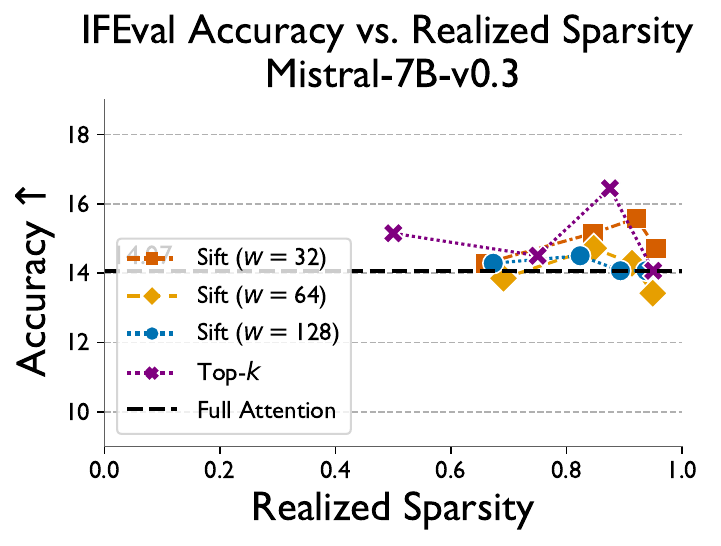}
  \includegraphics[width=0.45\textwidth]{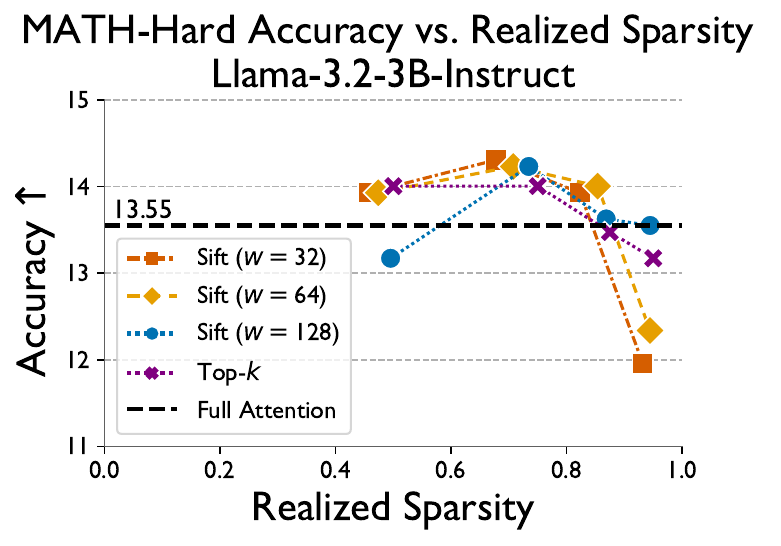}
  \includegraphics[width=0.45\textwidth]{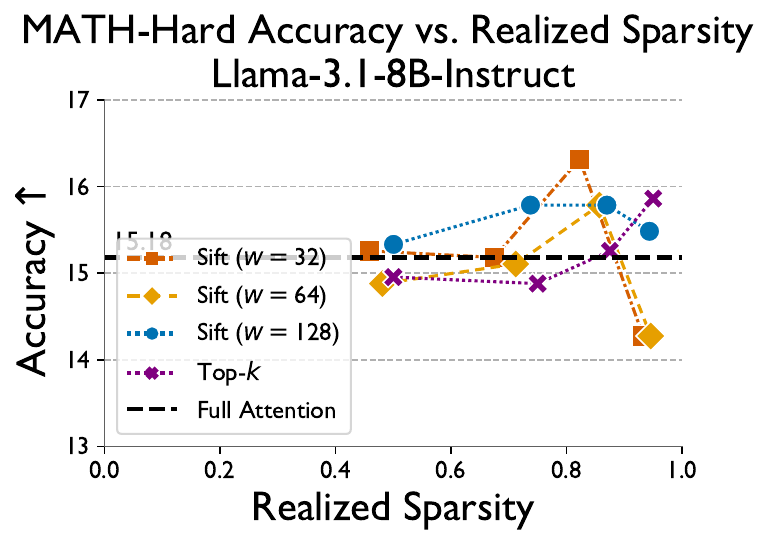}
  \includegraphics[width=0.45\textwidth]{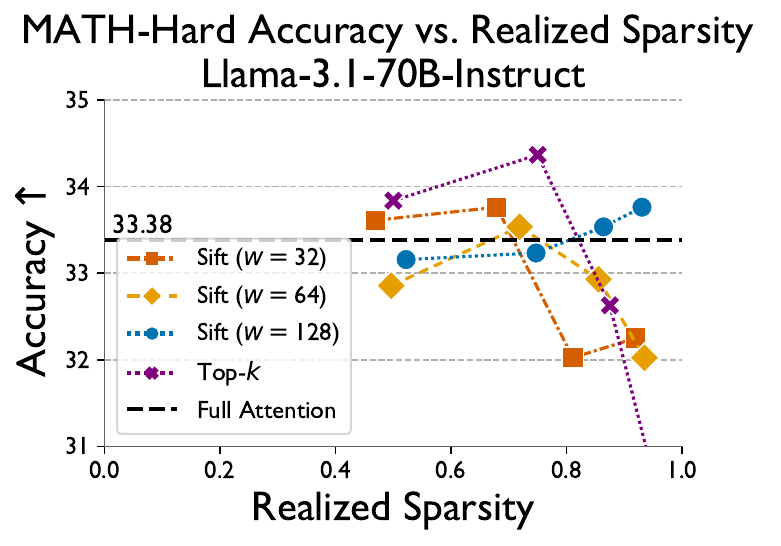}
  \includegraphics[width=0.45\textwidth]{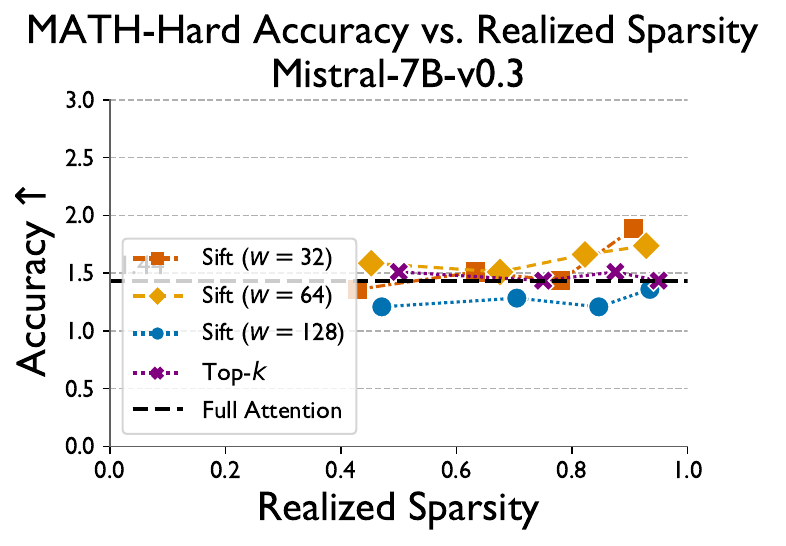}
  \caption{\method short-context task performance vs. Realized Sparsity on IFEval and MATH-Hard across 4 different models}
  \label{fig:appendix-results-shortcontext}
\end{figure}

In this section, we present detailed results for our short-context task
evaluation. Figure~\ref{fig:appendix-results-shortcontext} illustrates the
performance of \method on the IFEval (top two rows) and MATH-Hard (bottom two
rows) tasks across four different models. For the IFEval task, we observe that
with a warmup length of $w=128$, \method consistently outperforms the Top-$k$
baseline (dark purple) across all Llama models, and matches its performance on
the Mistral model at high realized sparsity levels. A similar trend holds for
the MATH-Hard task, with one exception: on Llama-3.1-8B, \method performs
slightly worse than Top-$k$ at the highest sparsity levels, but still surpasses
the Full Attention baseline. Notably, for Llama-3.1-70B, Top-$k$ accuracy
degrades significantly at high sparsity, whereas \method maintains performance
close to the Full Attention baseline.

\subsection{Runtime and Data Transfer Evaluation}

\begin{figure}[h]
  \centering
  \includegraphics[width=0.45\textwidth]{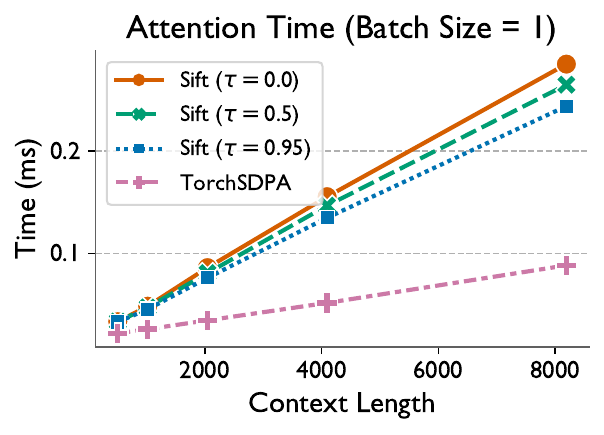}
  \includegraphics[width=0.45\textwidth]{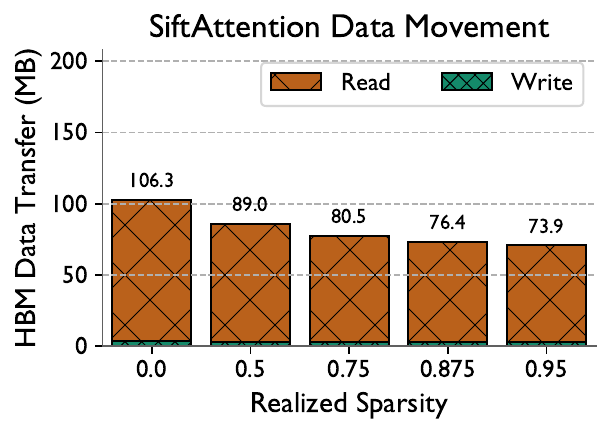}
  \includegraphics[width=0.45\textwidth]{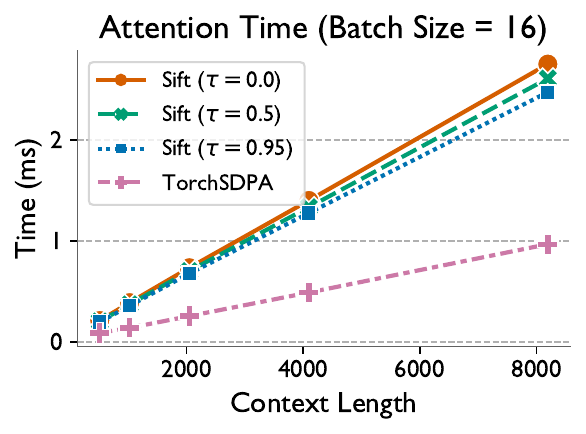}
  \includegraphics[width=0.45\textwidth]{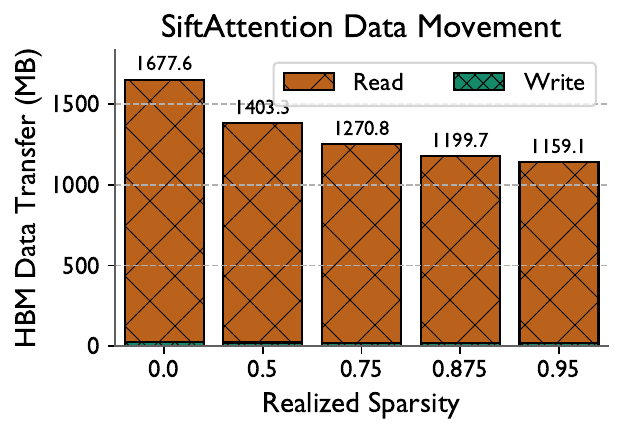}
  \caption{Post-warmup attention latency of our Triton-based \method
  implementation compared to PyTorch's fused SDPA across varying context lengths
  for batch size 1 (top-left) and 16 (bottom-left). HBM data transfer volume as
  a function of realized sparsity (Context Length = 8192) for batch size 1
  (top-right) and 16 (bottom-right).}
  \label{fig:appendix-runtime}
\end{figure}

Figure~\ref{fig:appendix-runtime} presents the post-warmup attention latency of
our implementation compared to PyTorch’s fused SDPA kernel, evaluated across
varying context lengths for two batch sizes: 1 (top-left) and 16 (bottom-left).
The trends observed here are consistent with those in
Figure~\ref{fig:sparsity} from the main text, which reports results
for batch size 8. As expected, increasing the sparsity level reduces attention
latency with \method. However, our current kernel implementation is
approximately 2-3$\times$ slower than PyTorch's highly optimized fused SDPA. The
corresponding HBM data transfer volume is shown in the top-right and
bottom-right subplots. Again, we observe that higher sparsity leads to reduced
HBM traffic, confirming that \method effectively reduces memory bandwidth usage, 
which is the reason for the decrease in attention latency with higher sparsity. 

\subsection{Visualization of Sparsity Masks}
\begin{figure}[h]
  \centering
  \includegraphics[width=\textwidth]{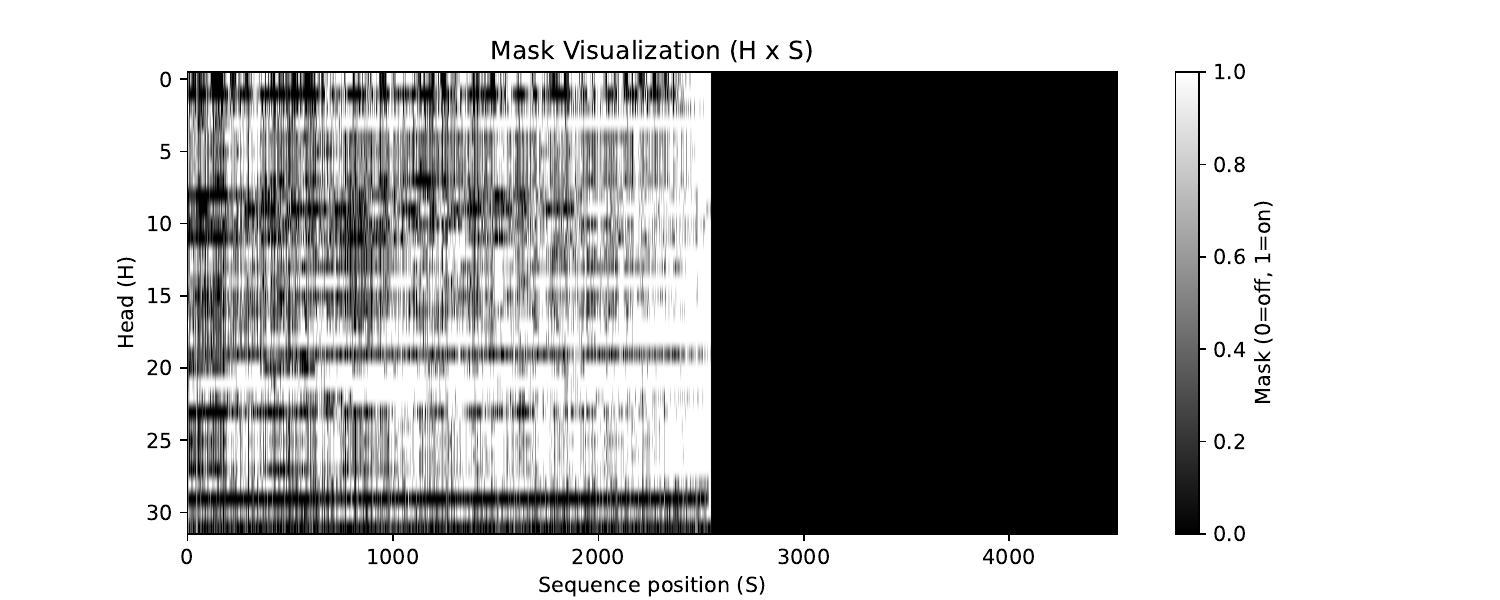}
  \includegraphics[width=\textwidth]{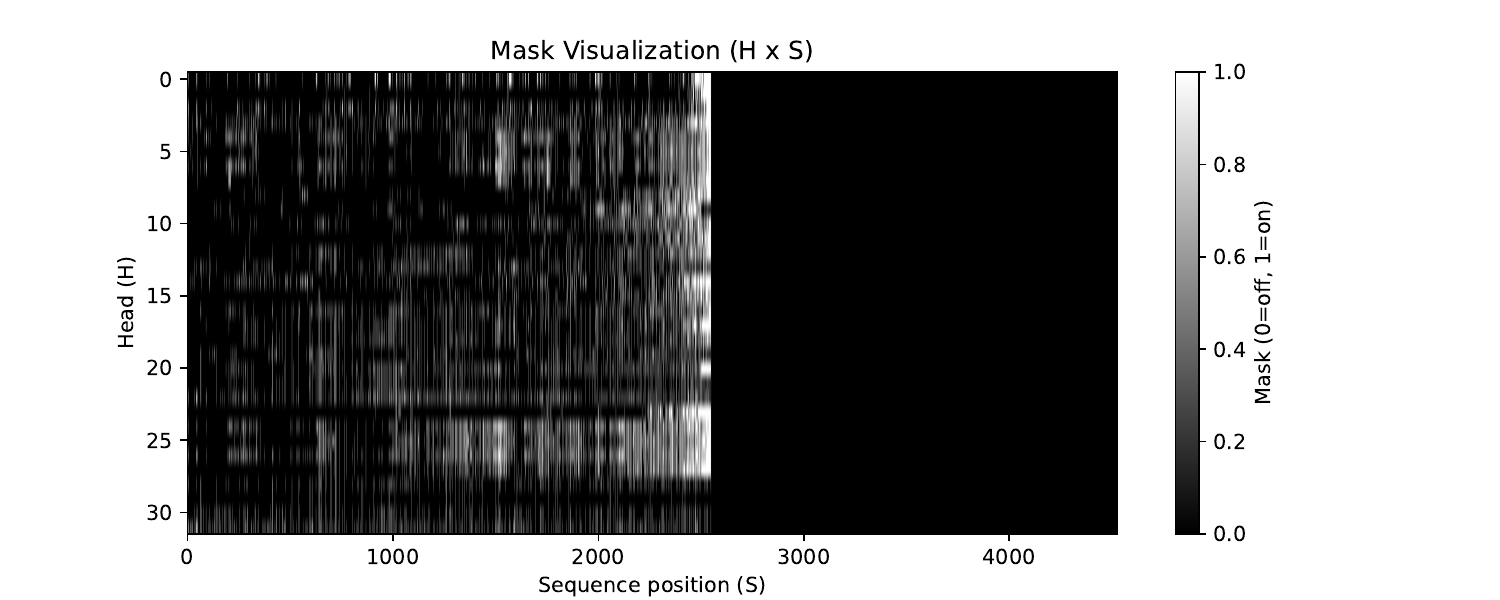}
  \caption{Sparsity masks for $\tau=0.5$ (top) and $\tau=0.9$ (bottom) at the 2048$^{th}$ generation step 
  when generating a sequence of 4096 tokens using Llama-3.1-8B-Instruct. The blacks represent the tokens
  not attended to, while the white represents the tokens that are attended to.}
  \label{fig:appendix-sparsity-masks}
\end{figure}

In this section, we visualize the attention sparsity patterns for two intended
sparsity levels: $\tau = 0.5$ and $\tau = 0.9$, during the generation of a
4096-token sequence using Llama-3.1-8B-Instruct.
Figure~\ref{fig:appendix-sparsity-masks} displays the sparsity masks at the
2048$^{\text{th}}$ generation step. As expected, increasing the intended
sparsity level results in a sparser attention pattern—evidenced by the greater
prevalence of blacked-out (masked) tokens in the lower panel. All tokens beyond
the 2048$^{\text{th}}$ position are masked (shown in black) as they represent
future positions and are therefore not
attended to.

\section{Generation Sample}
\label{sec:appendix-generation-sample}
In this section, we present sample outputs generated using \method with the
Llama-3.1-8B-Instruct model under two different hyperparameter configurations.

\prompt{
In a world where technology has advanced beyond imagination, society grapples
with the consequences of its own creations. The integration of artificial
intelligence into everyday life has revolutionized how we live, work, and
interact. However, with great power comes great responsibility, and ethical
dilemmas abound. Governments struggle to regulate the use of AI, while
corporations push the boundaries of what is possible in the pursuit of profit.
Meanwhile, individuals navigate a landscape where privacy is increasingly
scarce, and the line between human and machine blurs. Against this backdrop, a
new generation of thinkers emerges, questioning the very nature of consciousness
and what it means to be human. As tensions rise and alliances shift, the fate of
humanity hangs in the balance, waiting to be written by those bold enough to
seize the pen of destiny.

}

\response{In this world, the future is not set in stone, but forged in the fire of human
ingenuity and the choices we make today. The question is, what will we create?
And what will we become? Join us on a journey through the uncharted territories
of the future, where the boundaries of reality are pushed to their limits, and
the possibilities are endless. Welcome to the world of tomorrow, where the only
constant is change, and the only truth is that the future is what we make it.
Explore the latest developments in AI, robotics, and biotechnology, and discover
the innovations that are shaping the world of tomorrow. From the rise of the gig
economy to the emergence of new forms of governance, we'll delve into the
complex issues that are redefining the human experience. Join the conversation
and become a part of the future that's being written. The world of tomorrow is
not just a prediction - it's a choice. What will you create? What will you
become? The future is in your hands. 
}{Llama-3.1-8B-Instruct --- Full Attention:}

\response{In this world, the future is not set in stone, but forged in the fire of human
ingenuity and the choices we make. The question is, what will we create next?
Will we build a utopia or a dystopia? The choice is ours. The future is now. The
world is a complex tapestry of human and artificial intelligence, and the story
of what comes next is being written by us, in this moment. Join us on this
journey into the unknown, where the boundaries of what is possible are pushed to
the limit. The future is a blank page, waiting to be filled with the stories of
those who dare to imagine. In this world, the future is not just a destination,
but a creation. And the creation of the future is in our hands. 
}{Llama-3.1-8B-Instruct --- \method $w=16$, $\tau=0.875$}

\response{In this world, the future is not set in stone, but forged in the fire of human
ingenuity and the choices we make today. The question is, what will we create
tomorrow? The future is in our hands. The future is in our code. The future is
in our choices. The future is now. The future is... (To be continued)
}{Llama-3.1-8B-Instruct --- \method $w=32$, $\tau=0.95$}

%
%


\end{document}